\documentclass[journal]{IEEEtran}

\usepackage[utf8]{inputenc}
\usepackage[framemethod=tikz]{mdframed}
\usepackage{xcolor}
\usepackage{enumitem}
\usepackage{caption}
\usepackage{subcaption}
\usepackage{url}
\usepackage{amsmath,amssymb,amsthm,amsfonts}
\usepackage{algorithm}
\usepackage[noend]{algpseudocode}
\usepackage{hyperref}
\usepackage{graphicx}
\usepackage{textcomp}
\usepackage{comment}
\usepackage{dsfont}
\def\BibTeX{{\rm B\kern-.05em{\sc i\kern-.025em b}\kern-.08em
    T\kern-.1667em\lower.7ex\hbox{E}\kern-.125emX}}
    
\newcommand{\E}{\mathbb{E}}
\newcommand{\D}{\widehat{D}}
\newcommand{\g}{\gamma}
\newcommand{\tg}{\widetilde{\gamma}}
\newcommand{\hg}{\widehat{\gamma}}
\newcommand{\ind}[1]{\mathds{1}\{#1\}}
\newtheorem{remark}{Remark}
\newtheorem{theorem}{Theorem}
\newtheorem{definition}{Definition}
\newtheorem{lemma}{Lemma}

\begin{document}
\title{Learning Latent Events from Network Message Logs}
\author{Siddhartha~Satpathi,
        Supratim~Deb,
        R Srikant,
        and He Yan
\thanks{The work of Satpathi and Srikant was partially supported by a grant from AT\&T, NSF Grant NeTS 1718203 and NSF CPS ECCS 1739189.}

\thanks{S. Satpathi and R. Srikant are with UIUC. S. Deb and H. Yan are with AT\&T labs}}
\maketitle
\begin{abstract}
We consider the problem of separating error messages generated in large distributed data center networks into error events. In such networks, each error event leads to a stream of messages generated by hardware and software components affected by the event. These messages are stored in a giant message log. We consider the unsupervised learning problem of identifying the signatures of events that generated these messages; here, the signature of an error event refers to the mixture of messages generated by the event. One of the main contributions of the paper is a novel mapping of our problem which transforms it into a problem of topic discovery in documents. Events in our problem correspond to topics and messages in our problem correspond to words in the topic discovery problem. However, there is no direct analog of documents. Therefore, we use a non-parametric change-point detection algorithm, which has linear computational complexity in the number of messages, to divide the message log into smaller subsets called episodes, which serve as the equivalents of documents. After this mapping has been done, we use a well-known algorithm for topic discovery, called LDA, to solve our problem. We theoretically analyze the change-point detection algorithm, and show that it is consistent and has low sample complexity. We also demonstrate the scalability of our algorithm on a real data set consisting of $97$ million messages collected over a period of $15$ days, from a distributed data center network which supports the operations of a large wireless service provider.
\end{abstract}


\section{Introduction}
The delivery of modern data and web-based services requires the execution of a chain of network functions at different elements in distributed data-centers. This is true for video-based services, gaming services, cellular data/voice services, etc., each of which requires processing from multiple coupled networked entities hosting different network functions. For example, modern wireless networks rely on servers and virtual machines (VM) residing in distributed data centers to establish voice calls or data sessions, authenticate users, check user compliance with monthly voice/data limits, verify if users have paid their monthly bills, add to users' bills for extra services, etc., all of which are done before completing a call. Efficient management and operations of these services is of paramount importance as networks grow increasingly complex with the advent of technologies like virtualization and 5G. An integral component of network management is the ability to identify and understand \emph{error events}, when failures occur in the hardware and/or software components of the network. However, the complex interdependence between coupled networking functions poses a significant challenge in characterizing an error event due to the fact that error messages can be generated in network elements beyond the actual source of error. In this paper, we are interested in the problem of mining latent error event information from messages generated by servers, VMs, base stations, routers, and links in large-scale distributed data center networks. The mined events are useful for troubleshooting purposes. Also, the correlations captured through each learned event could be subsequently used for on-line detection of potential errors. While our methodology is broadly applicable to any type of data center network, we validate our algorithms by applying them to a large data set provided by a major wireless network service provider, so we will occasionally use terminology specific to this application to motivate our problem and solution methodology.

In most  operational networks, all messages and alarms from distributed network elements are logged with time stamps into message logs. The logs from different network elements could be pooled together in a central database for subsequent analysis. 
While mining error logs have been studied extensively in different contexts, (see~\cite{lishwartz:kdd2017:tutorial, Li:2017:DTC:3101309.3092697} for excellent surveys; also see Section~\ref{sec:related}), there are some fundamental differences in our setting. Modern data center and communication networks consist of components bought from different vendors, and each component is designed to generate an error message when it cannot execute a job. This poses a challenge in mining messages because there is no common model or standard that dictates the content and format of these error messages. Another challenge stems from the fact that each end-to-end service consists of multiple network functions each of which generates diverse error messages when failures happen. The following example provides an illustration.

	{\bf Motivating Example:} Suppose Alice makes a cellphone call to Bob. This call is first routed through a base station which is attached to a data center verifying the caller credentials. If Alice is not at her home location, a VM at this data center must contact a database at her home location to verify her credentials. Once the credentials are verified, the caller's cellular base station connects to the base station near Bob through a complicated network spanning many geographical locations. Consider two potential error scenarios: (i) an error occurs at a router in the path from  Bob to Alice's base station, (ii) error at a router connecting the data centers verifying the caller's credentials. In either scenario, the call will fail to be established leading to the generation of error messages not only at the failed routers but also at network functions (implemented in a cluster of VMs) responsible for call establishment. Furthermore, if the error leads to additional call failures, then respective base-stations could send alarms indicating higher than normal call failures. Additionally, depending on vendor of a given network element, the timing and content of the error messages could be different.

Indeed, the source, timing, and message-components of the error are all latent. In this paper we are interested in extracting patterns from messages generated by common faults/errors (also referred to as events). Specifically, our goal of this paper is to mine event signatures (i.e., distribution of messages for each event)  and event occurrences (i.e., the begin and the end time of each event) from the message log.  Based on the motivating example, we now note the following fundamental characteristics which make our error event mining problem challenging:
\begin{itemize}[leftmargin=10pt]
	\item  In our setting, the source of an error is usually not known. There could {be} error messages due to network-component level failures or due to network service-level failure. In case of a service-level failure, error-message could be generated by a component that is functional by itself. For example, when the link between an authentication server and the network core fails, this could lead to call establishment failures which are logged by network functions responsible for call establishment. Furthermore, the same type of error log-message could be generated due to many different errors. From a data modeling point of view, each (latent) event can be viewed as a probabilistic-mixture of multiple log-messages at different elements and also, the set of log messages generated by different events could have non-zero intersection.
	
	\item Each error event can produce a sequence of messages, including the same type of message multiple times, and the temporal order between distinct messages from the same event could vary based on the latency between network elements, network-load, co-occurrence of other uncorrelated events, etc.  Thus, the temporal pattern of messages may also contain useful information for our purpose. In our model, the message occurrence times are modeled as a stochastic process.
	
	\item These messages could correspond to multiple simultaneous events without any further information on the start-time and end-time of each event.
	
	\item An additional challenge arises from the fact that network topology information is unknown, because modern networks are very complicated and are constantly evolving due to the churn (addition or deletion) of routers and switches.  Third-party vendor software and hardware have no way of providing information to localize and understand the errors. Thus, topological information cannot be used for event mining purposes.
\end{itemize}
The practical novelty of our work comes from modeling  for all of the above factors and proposing scalable algorithms that learn the latent event signatures (the notion of signature will be made precise later) along with their occurrence times.  

\begin{remark}
	\label{rem:event}
	\vspace{-0.1in}
	{\em We note that, in different works on event mining (see Section~\ref{sec:related}), the concept of event is different depending on the problem-context. It could either mean semantic-event or message template, or a cluster of such templates, or in some cases event itself is equivalent to message (where tagged event streams are available) or a transaction/system-event. In our work, an {\em event} simply refers to a real-world occurrence of { a fault/incident} somewhere in the distributed/networking system such that each event leads to a generation of error messages at multiple network elements. }
\end{remark}

\subsection{Contributions}

We model each event as a probabilistic mixture of messages from different sources \footnote{It is more precise to use the terminology event-class to refer to a specific fault-type; each occurrence can be referred to as an instance of some event class. However, for simplicity, we simply refer to event-class as event and we just say occurrence of the event to mean instance of this class.}. In other words, the probability distribution over messages characterizes an event, and thus acts as the signature of the event. Each occurrence of an event also has a start/end time and several messages can be generated during the occurrence of an event. We only observe the messages and their time-stamps while the event signatures and duration window is unknown; also there could be multiple simultaneous events occurring in the network. Given this setting, we study the following unsupervised learning problem: {\em given collection of time-stamped log-messages, learn the latent event signatures and event start/end times.}

The main contributions of the paper are as follows:
\begin{itemize}[leftmargin=10pt]
    \vspace{-0.05in}
    \item {\em Novel algorithmic framework:} We present a novel way of decomposing the problem into simpler sub-problems. Our method, which we will call CD-LDA, decomposes the problem  into two parts: the first part consists of a change-point detection algorithm which identifies time instants at which either a new event starts in the network or an existing event comes to an end, and, the second part of the algorithm uses {Latent Dirichlet Allocation} (LDA) (see \cite{LDA}) to classify messages into events. This observation that one can use change-point detection, followed by LDA, for event classification is one of the novel ideas in the paper.
    
    \item {\em Scalable change-point detection:} While the details of the LDA algorithm itself are standard,  non-parametric change-point detection as we have used in this paper is not as well studied. We adapt an idea from \cite{nonparamchpt} to design an $O(n)$ algorithm where $n$ is the number of messages in the message log. Our change detection algorithm uses an easy to compute total-variation (TV)  distance. 
    We analyze the sample complexity of (i.e, the number of samples required to detect change points with a high-degree of accuracy) of our change-point detection algorithm using the method of types and Pinsker's inequality from information theory. To the best of our knowledge, no such sample complexity results exist for the algorithm in \cite{nonparamchpt}.
    
    \item {\em Experimental validation:}  We use two different real-world data sets from a large operational network to perform the following validation of our approach. First, we compare our algorithm to two existing approaches adapted to our setting: a Bayesian inference-based algorithm and Graph-based clustering algorithm.  We show the benefits of our approach compared to these methods in terms of scalability and performance, by applying it to small samples extracted from a large data set consisting of $97$ million messages. Second, we validate our method against two real world events by comparing the event signature learned by our method with domain expert validated event signature for a smaller data set consisting of $700$K messages\footnote{Note that manual inference of event signatures is not scalable; we did this for the purpose of validation.}. 
    Finally, we also show results to indicate scalability of our method by applying to the entire $97$ million message data set.
    
\end{itemize}
We note that this paper is {an extended version of} \cite{KDDworkshop} that appeared in a workshop.

\subsection{Context and Related Work}
\label{sec:related}


Data-driven techniques have been shown to be very useful in extracting meaningful information out of system-logs and alarms for large and complex systems. The primary goal of this ``knowledge" extraction is to assist in diagnosing the underlying problems responsible for log-messages and events. Two excellent resources for the large body of work done in the area are~\cite{lishwartz:kdd2017:tutorial, Li:2017:DTC:3101309.3092697}.  
Next, we outline some of the key challenges in this knowledge extraction, associated research in the area, and our problem in the context of existing work. 

	 {\bf Mining and clustering unstructured logs:} Log-messages are unstructured textual data without any annotation for the underlying fault. A significant amount of research has focused on converting unstructured logs to common {\em semantic events}~\cite{Li:2017:DTC:3101309.3092697}. Note that the notion of {\em semantic events} is different from the actual real-world events responsible for generating the messages, nevertheless, such a conversion helps in providing a canonical description of the log-messages that enables subsequent correlation analysis. These works exploit the structural similarity among different messages to either compute an intelligent log-parser or cluster the messages based on message texts~\cite{Makanju:2009:CEL:1557019.1557154,Li:2005:IFM:1081870.1081972,DBLP:conf/icdm/TangL10,Li:2017:DTC:3101309.3092697}. Each cluster can be viewed as an semantic event which can help in diagnosing the underlying root-cause. One work closely related to ours is \cite{newrelated}, in which the authors mine network log messages to first extract templates and then learn pairwise \textit{implication} rules between template-pairs. 
	Our setting and objective are somewhat different, we model events as message-distributions from different elements with  each event occurrence having certain start and end times;  the messages belonging to an event and the associated occurrence time-windows are hidden (to be learned). A more recent work~\cite{Wu:2017:SED:3097983.3098124} develops algorithms to mine underlying structural-event as a work-flow graph. The main differences are that, each transaction is a fixed sequence of messages unlike our setting where each message could be generated multiple times based on some hidden stochastic process, and furthermore, in our setting, there could be multiple events manifested in the centralized log-server.
	
	{\bf Mining temporal patterns:} Log-messages are time-series data and thus the temporal patterns  contain useful information.  Considerable amount of research has gone into learning latent patterns, trends and relationship between events based on timing information in the messages~\cite{Agrawal:1995:MSP:645480.655281, Cheng:2014:FSE:2623330.2623709,bigdataconf-ZengWWLS16}. We refer to~\cite{Mooney:2013:SPM:2431211.2431218, Li:2017:DTC:3101309.3092697,Silva:2013:DSC:2522968.2522981} for survey of these approaches. Extracted event-patterns could be used to construct event correlation graphs that could be mined using  techniques such as graph-clustering. Specifically, these approaches are useful  when event-streams are available as time-series. We are interested in scenarios where each event is manifested in terms of time-series of unstructured messages and furthermore, same message could arise from multiple events. Nevertheless, certain techniques developed for temporal event mining could be adapted to our setting as we describe in Section~\ref{sec:algC}; our results indicate that such an adaptation works well under certain conditions. Note that, our goal is to also learn the event-occurrence times.

	{\bf Event-summarization:}  In large dynamic systems, messages could be generated from multiple components due to reasons ranging from software bugs, system faults, operational activities, security alerts etc. Thus it is very useful to have a global summarized snapshot of messages based on logs. Most works in this area exploit the inter-arrival distribution and co-occurrence of events~\cite{Jiang:2011:NES:2063576.2063688, Wang:2010:AAE:1807167.1807189, Peng:2007:ESS:1281192.1281305, Tatti:2012:LSS:2339530.2339606, Li:2017:DTC:3101309.3092697} to produce summarized correlation between events. These methods are useful when the event-stream is available and possible event-types are known in advance. This limits the applicability to large-systems like ours where event types are unknown along with their generation time-window.


The body of work closest to out work are the works on event-summarization. However, there are some fundamental differences in our system: (i) we do not have a readily available event-stream, instead, our observables are log-messages, (ii) the event-types are latent variables not known in advance and all we observe are message streams, (iii) the time-boundaries of different latent-events is a learning objective, and (iv) since we are dealing with large system with multiple components where different fault-types are correlated, the same message could be generated for different root-causes (real-world events). 

Apart from the above, a recent paper \cite{deeplog} which uses deep learning models for anomaly detection in message logs by modelling logs as a natural language sequence is also worth a mention.


\section{Problem Statement and Preliminaries}
Before we describe our problem statement, we first explain the notion of messages in the context of our work.

{\bf Message:}  In our work, messages generated by different network elements are one of two types: {\em syslog texts} in the form of raw-texts, and {\em alarms}.
\begin{enumerate}[leftmargin=15pt]
	\item {\em Syslog texts:} These are raw-textual messages sent by software components from different elements to a logging server. Raw syslog data fields include timestamp, source, and message text. Since the number of distinct messages are very large and many of them have common patterns, it is often useful~\cite{Makanju:2009:CEL:1557019.1557154,Li:2005:IFM:1081870.1081972,DBLP:conf/icdm/TangL10,Li:2017:DTC:3101309.3092697} to decompose the message text into two parts: an {\em invariant} part called template, and {\em parameters} associated with  template. For example, a syslog message 
	``\texttt{service wqffv failed due to connection failure to IP address a.b.c.d using port 8231}" would reduce to template ``\texttt{service wqffv failed due to connection failure to IP address * using port *}."  There are many existing methods to extract such templates\cite{lishwartz:kdd2017:tutorial, Li:2017:DTC:3101309.3092697}, ranging from tree-based methods to NLP based methods.  In our work, we have a template-extraction pre-processing step before applying our methods. We also say {\em message} to simply mean the extracted templates.
	
	\item {\em Alarms:}  Network alarms are indication of faults and each alarm type refers to the specific fault condition in a network element. Each alarm has a unique name and the occurrences are also tagged with timestamps. In this work, we view each alarm as a message. Note that, since each alarm has a unique name/id associated with it, we do not pre-process alarms before applying our methods. Example of alarms are \texttt{ mmscRunTImeError, mmscEAIFUnavailable}  sent from a network service named MMSC.
\end{enumerate}

\begin{figure}[ht]
	\vspace{-0.1in}
	\centerline{\includegraphics[scale = 0.35]{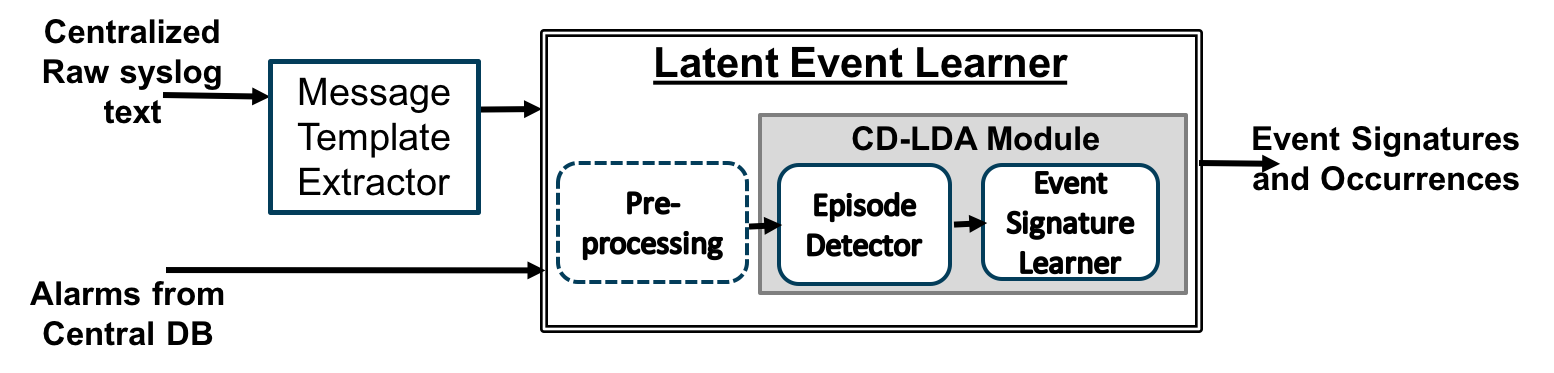}}
	\caption{Figure showing the  machine-learning pipeline. Our main contribution is in  "Latent Event Learner" module, specifically proposing the CD-LDA algorithm.}
	\label{fig:mlpipeline}
\end{figure}
{\bf Problem Statement:} We are given a data set $\mathcal{D}$ consisting of messages generated by error events in a large distributed data-center network. We assume that the messages are generated in the time interval $[0,T].$ The set of messages in the data set come from discrete and finite set $\mathcal{M}.$ 

We use the term message to mean either a template extracted from a message or an alarm-id.
Each message has a timestamp associated with it, which indicates when the message was generated. Suppose that an event $e$ started occurring at time $S_e$ and finished at time $F_e.$ In the interval of time $[S_e,F_e],$ event $e$ will generate a mixture of messages from a subset of $\mathcal{M},$ which we will denote by $\mathcal{M}_e.$ In general, an event can occur multiple times in the data set. If an event $e$ occurs multiple times in the data set, then each occurrence of the event will have start and finish times associated with it.

As noted before, for simplicity, we will say event to mean an event-class and occurrence of an event to mean an instance from the class. An event $e$ is characterized by its message set $\mathcal{M}_e$ and the probability distribution with which messages are chosen from $\mathcal{M}_e,$ which we will denote by $p^{(e)},$ i.e., $p^{(e)}_m$ denotes the probability that event $e$ will generate a message $m\in\mathcal{M}_e.$ For compactness of notation, one can simply think of $p^{(e)}$ as being defined over the entire set of messages $\mathcal{M},$ with $p_m^{(e)}=0$ if $m\notin \mathcal{M}_e.$ Thus, $p^{(e)}$ fully characterizes the event $e$ and can be viewed as the signature of the event. We assume that the support set of messages for two different events are not identical.

It is important to note that the data set simply consists of messages from the set $\mathcal{M};$ there is no explicit information about the events in the data set, i.e., the event information is latent. The goal of the paper is to solve the following inference problem: from the given data set $\mathcal{D},$ identify the set of events that generated the messages in the data set, and for each instance of event, identify when it started and finished. In other words, the output of the inference algorithm should contain the following information:
\begin{itemize}
    \item The number of $E$ events which generated the data set.
    \item The signatures of these events: $p^{(1)}, p^{(2)},\ldots,p^{(E)}.$
    \item For each event $e\in\{1,2,\ldots,E\},$ the number of times it occurred in the data set and, for each occurrence, its start and finish times.
\end{itemize} 

\begin{sloppypar}
\textbf{Notations:} We use the notation $X_i\in \mathcal{M},$ for the $i^{th}$ message. Also, let $t_i$ be the timestamp associated with the $i^{th}$ message. 
Thus the data set $\mathcal{D}$ can be characterized by tuples $(X_1,t_1), (X_2,t_2), \ldots (X_n,t_n)$ of $n$ data points.  
\end{sloppypar}

{\bf Machine-learning pipeline:} In Figure~\ref{fig:mlpipeline}, we show the machine-learning pipeline for completeness. This paper focuses on the module ``Latent Event Learner" which has data-processing step followed by the key proposed algorithm in the paper, namely CD-LDA algorithm which we describe in Section~\ref{sec:cdlda}. Syslog texts require more pre-processing while alarms do not. We have shown the two types of messages in the pipeline figure, but for the purposes for developing an algorithm, in the rest of the paper, we only refer to messages without distinguishing between them. 

\section{Algorithm CD-LDA}
\label{sec:cdlda}
We now present our solution to this problem which we call CD-LDA (Change-point Detection-{Latent Dirichlet Allocation}). The key novelty in the paper is the connection that we identify between event identification in our problem and topic modeling in large document data sets, a problem that has been widely studied in the natural language processing literature. In particular, we process our data set into a form that allows us to use a widely-used algorithm called LDA to solve our problem. In standard LDA, we are given multiple documents, with many words in each document. The goal is to identify the mixture of latent topics that generated the documents, where each topic is identified with a collection of words and a probability distribution over the words. Our data set has similar features: we have events (which are the equivalents of topics) and messages (which are the equivalents of words) which are generated by the events. However, we do not have a concept of documents. A key idea in our paper is to divide the data set into smaller data sets, each of which will be called an episode. The episodes will be the equivalents of documents in our problem. We do this using a technique called non-parametric change-point detection. 

Now we describe the concept of an episode.
An episode is an interval of time over which the same set of events occur i.e. there is no event-churn, and at time instants on either side of the interval, the set of events that occur are different from the set of events in the episode. Thus, we can divide our data set of events such that no two consecutive episodes have the same set of events. We present an example to clarify the concept of an episode. Suppose the duration of the message data set $T=10.$ Suppose event $1$ occurred from time $0$ to time $5,$ event $2$ occurred from time $4$ to time $8,$ and event $3$ occurred from time $5$ to time $10.$ Then there are four episodes in this data set: one in the time interval $[0,4]$ where only one event occurs, one in the time interval $[4,5]$ where events $1, 2$ occur, one in the time interval $[5,8]$ where events $2, 3$ occur and finally one in $[8,10]$ where only event $3$ occurs. We assume that between successive episodes, at most one new event starts or one existing event ends.

We use change-point detection to identify episodes. To understand how the change-point detection algorithm works, we first summarize the characteristics of an episode:
\begin{itemize}
    \item An episode consists of a mixture of events, and each event consists of a mixture of messages.
    \item Since neighboring episodes consist of different mixtures of events, neighboring episodes also contain different mixtures of messages (due to our assumption that different events do not generate the same set of messages).
    \item Thus, successive episodes contain different message distributions and therefore, the time instances where these distributions change are the episode boundaries, which we will call \emph{change points}.
    \item In our data set, the messages contain time stamps. In general, the inter-arrival time distributions of messages are different in successive episodes, due to the fact that the episodes represent different mixtures of events. This fact can be further exploited to improve the identification of change points.
\end{itemize}

Based on our discussion so far in this section, CD-LDA has two-phases as follows:
\begin{enumerate}[leftmargin=15pt]
	\item {\em Change-point detection:} In this phase, we detect the start and end time of each episode. In other words, we identify the time-points where a new event started or an existing event ended. This phase is described in detail in Section~\ref{subsec:CD}.
	
	\item{\em Applying LDA:} In this phase, we show that, once  episodes are known,  LDA based techniques can be used to solve the problem of computing message distribution for each event. Subsequently, we can also infer the occurrence times for each event. This phase along with the complete algorithm is described in Section~\ref{sec:LDA}.
\end{enumerate}

\subsection{Change-point Detection}
\label{subsec:CD}
Suppose we have $n$ data points and a known number of change points $k$. The data points between two consecutive change points are drawn i.i.d from the same distribution\footnote{The i.i.d. assumption is
not always true in practice as messages could be sparser in time in the beginning of an event. Indeed, the algorithms developed in this work does not rely on the i.i.d. assumption, however, the assumption allows
us to prove useful theoretical guarantees}. In the inference problem, each data point could be a possible change point. A naive exhaustive search to find the $k$ best locations would have a computational complexity of $O(n^k)$. Nonparametric  approaches to change-point detection aim to solve this problem with much lower complexity even when the number of change points is unknown and there are few assumptions on the family of distributions, \cite{nonparamchpt1}, \cite{nonparamchpt},\cite{nonparamchpt2}. 

The change point detection algorithm we use is hierarchical in nature. This is inspired by the work in \cite{nonparamchpt}. Nevertheless our algorithm has certain key differences as discussed in section \ref{sec:complexity}. It is easier to understand the algorithm in the setting of only one change point, i.e., two episodes. Suppose that $\tau$ is a candidate change point among the $n$ points. The idea is to measure the change in distribution between the points to the left and right of $\tau$. {We use the TV distance between the empirical distributions estimated from the points to the left and right of the candidate change point $\tau$. In our context the TV distance between two probability mas functions $p$ and $q$ is given by one half the $L1$ distance $0.5||p-q||_1$.} This is maximized over all values of $\tau$ to estimate the location of the change point. If the distributions are sufficiently different in the two episodes the TV distance between the empirical distributions is expected to be highest for the correct location of the change point in comparison to any other candidate point $\tau$ (we rigorously prove this in the proof Theorem \ref{lemma:consistency}, \ref{th:multiplechptSmaple}). 

Further, we also have different inter-arrival times for messages in different episodes. Hence we use a combination of TV distance and mean inter-arrival time as the metric to differentiate the two distributions\footnote{One can potentially use a weighted combination of the TV distance and mean inter-arrival time as a metric with the weight being an hyper parameter. While the unweighted metric performs well in out real-life datasets, it is an interesting future direction of research to understand how to optimally choose a weighted combination in general.} We denote this metric by $\D(l)$.


\begin{align}
    \widehat{D}(l) = \|\widehat{p}_L(l) - \widehat{p}_R(l)\|_1 + |\widehat{\E} S_L(l) - \widehat{\E} S_R(l) |,
    \label{eq:Dhat}
\end{align}
where $\widehat{p}_L(l),$ $\widehat{p}_R(l)$ are empirical estimates of message distributions to the left and right  $l$ and $\widehat{\E} S_L(l),$ $\widehat{\E} S_R(l)$ are empirical estimates of the mean inter-arrival time to the left and right of $l,$ respectively. 
The empirical distributions $\widehat{p}_L(l)$, $\widehat{p}_R(l)$ have $M$ components. For each $m\in\mathcal{M}$, we can write 
\begin{align}
    \widehat{p}_{L,m}(l) &= \dfrac{\sum_{i=1}^{l-1}\ind{X_i = m}}{l}\\
    \widehat{p}_{R,m}(l) &= \dfrac{\sum_{i=l}^{n}\ind{X_i = m}}{n - l}.\label{eq:defp}
\end{align}
 The mean inter-arrival time $\widehat{\E} S_L(l)$ and $\widehat{\E} S_L(l)$ are defined as
\begin{align}
     \widehat{\E} S_L(l) &= \dfrac{\sum_{i=1}^{l-1}\Delta t_i}{l}\\
     \widehat{\E} S_R(l) &= \dfrac{\sum_{i=l}^n\Delta t_i}{n-l}.\label{eq:defS}
\end{align}
We sometimes write $\D(l)$ as $\D(\tg n)$, where  the argument $l=\tg n$. Symbol $\tg$ denotes the index $l$ as a fraction of $n$ and it can take $n$ discrete values between $0$ to $1$. {$\mathds{1}\{A\}$ takes value $1$ only when event $A$ occurs and $0$ otherwise.}

Algorithm \ref{alg:onechangept} describes the algorithm in the one change point case. To make the algorithm more robust, we declare a change point only when the episode length is at least $\alpha n$ and the maximum value of the metric \eqref{eq:Dhat} is at least $\delta$.

\begin{figure}[ht]
\centerline{\includegraphics[scale = 0.3]{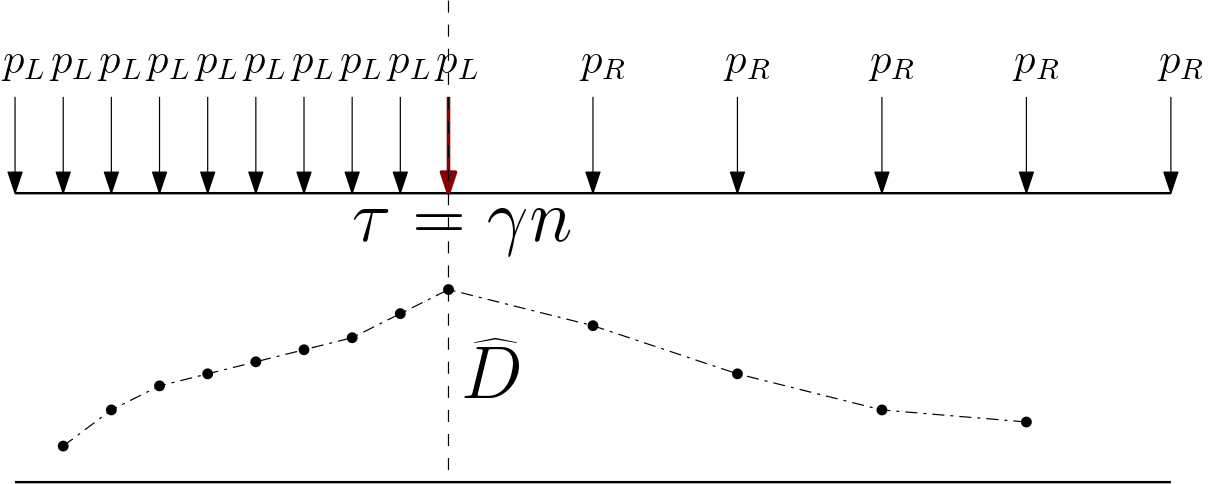}}
\caption{Example change point with two episodes}
\label{fig:example}
\end{figure}

Let us consider a simple example to illustrate the idea of change-point detection with one change-point. Suppose we have a sequence of messages with unequal inter-arrival times as shown in Fig. \ref{fig:example}. All the messages are the same, but the first half of the messages arrive at a rate higher than the second half of the messages. In this scenario, our metric reduces to the difference in the mean inter-arrival times between the two episodes. 
So, $\widehat{D} (l) = |\widehat{\E} S_L(l) - \widehat{\E} S_R(l) |$. The function $\widehat{D}$  in terms of data point $l$  for this example is shown in Fig \ref{fig:example}. As we show later in section \ref{sec:analysis}, the shape of $\widehat{D}$ will be close to the following when the number of samples is large: $\widehat{D}$ will be increasing to the left of change point $\tau = \gamma n$, attain its maximum at the change point and decrease to the right.

\begin{algorithm}
\begin{algorithmic}[1]

\State \textbf{Input}: parameter $\delta>0,\alpha > 0$.
\State \textbf{Output}: \textit{changept} denoting whether a change point exists and the location of the change point $\tau$.
\State Find $\tau \in \arg \max_{l} \D(l)$

\If {$\D(\tau)> \delta$ and $\alpha n <\tau<1 - \alpha n$} 
    \State \Return $changept = 1$, $\tau$.
\Else
    \State \Return $changept = 0$.
\EndIf

\caption{Change point detection with one change point}
\label{alg:onechangept}
\end{algorithmic}
\end{algorithm}



Next, we consider the case of multiple change points.
When we have multiple change points, we apply Algorithm \ref{alg:onechangept} hierarchically until we cannot find a change point. Algorithm \ref{alg:multiplechangept} \textsc{CD}$(\mathcal{D},\alpha, \delta)$ is presented below.

\begin{algorithm}
\begin{algorithmic}[1]
    
    \State \textbf{Input}: data points $\mathcal{D}$, minimum value of TV distance $\delta$, minimum episode length $\alpha$.
    
    \State \textbf{Output}: Change points $\tau_1,\ldots,\tau_k$.

    \State \textit{Run} \textsc{FindChangept}$(1,n)$.
    \Procedure{FindChangept}{$L,H$}
    
        \State \textit{changept}, $\tau \gets$ \textsc{Algorithm} \ref{alg:onechangept} ($X_L,X_{L+1},\ldots , X_H,\alpha,\delta$).
        
        \If{\textit{changept exists} }
            \State $\tau_l \gets$ \textsc{FindChangept}$(L,\tau)$, \State $\tau_h \gets$ \textsc{FindChangept}$(\tau,H)$.
            \State \Return $\{\tau_l,\tau,\tau_h\}$
        \Else
            \State \Return 
        \EndIf
    
    \EndProcedure
\end{algorithmic}
\caption{CD$(\mathcal{D},\alpha,\delta)$}
\label{alg:multiplechangept}
\end{algorithm}

The above algorithm tries to detect a single change point first, and if such a change point is found, it divides the data set into two parts, one consisting of messages to the left of the change point and the other consisting of messages to the right of the change point. The single change-point detection algorithm is now applied to each of the two smaller datasets. This is repeated recursively till no more change points are detected. 

\subsubsection{Discussion: What metric for change point detection?}
We have used the TV distance between two distributions to estimate the change point in metric \ref{eq:Dhat}. One can also use other distance measures like the $l_2$ distance, the Jensen-Shannon (J-S) distance, the Hellinger distance, or the metric used in \cite{nonparamchpt}. The metric used in \cite{nonparamchpt} is shown to be an unbiased estimator of the $l_2$ distance for categorical data in Appendix \ref{app:L2unbiased} of the supplementary material. We argue that for our data set, all of the above distances give similar performance. Our data set has 97m points and {39330} types of messages. In the region where the number of data points is much more than the dimension of the distribution, estimating a change point through all of the above metrics give order wise similar error rate. We show this through synthetic data experiments since we do not know the ground truth to compute the error in estimating the change point in the real dataset.

We present one such experiment with a synthetic dataset here. Consider two distributions $p$ and $q$ whose support set consists of $10$ points. We assume that $p$ is the uniform distribution, while $q[1]=q[2]=\ldots=q[5]=0.09$, and $q[6]=q[7]=\ldots=q[10]=0.11$.
There are $n = 25000$ data points. The first half of the data points are independently drawn from $p$ and the second half of the data points are drawn from $q$. Table \ref{tab:chptmetric} shows the absolute error in estimating the change point at $0.5n$ to be of the order of $10^{-2}$ for all the distance metrics.

\begin{table}
\begin{center}
\begin{tabular}{|c|c|c|c|c|c|}                     
    \hline
    \multicolumn{6}{|c|}{$||p - q||_1 = 0.1$} \\
    \hline
                                            Metric & $TV$ &  $l_2$ & Unbiased $l_2$, \cite{nonparamchpt} & J-S & Hellinger \\
    \hline
                                            $|\widehat{\tau}/n - 0.5|$& 0.021 & 0.030 & 0.025 & 0.030 & 0.030\\
    \hline
\end{tabular}
\end{center}

\caption{Comparison between different metrics for change point}
\label{tab:chptmetric}
\end{table}

We test the $l_1$ distance metric on real data and we show in section \ref{sec:visualValidation} that it is satisfactory. Since we do not know the ground truth, we take a small part of the real data set where we can can visually identify the approximate location of the major change points. The change point algorithm with $l_1$ metric correctly estimates these locations.

A graph based change point detection algorithm in \cite{chen2015} can be adapted to our problem such that the metric computation is linear in the number of messages. We can do this if we consider a graph with nodes as the messages and edges connecting message of the same type. But, one can show that the metric in \cite{chen2015} is not consistent for this adaptation.

\subsection{Latent Dirichlet Allocation}
\label{sec:LDA}

 In the problem considered in this paper, each episode can be thought of as a document and each message can be thought of as a word. Like in the LDA model where each topic is latent, in our problem, each event is latent and can be thought of as a distribution over messages. Unlike LDA-based document modeling, we have time-stamps associated with messages, which we have already used to extract episodes from our data set. 
 Additionally, this temporal information can also be used in a Bayesian inference formulation to extract events and their signatures. 
 However, to make the algorithm simple and computationally tractable, as in the original LDA model, we assume that there is no temporal ordering to the episodes or messages within the episodes. Our experiments suggest that this choice is reasonable and leads to very good practical results. However, one can potentially use the temporal information too as in \cite{tot, CTDM}, and this is left for future work.

If we apply the LDA algorithm to our episodes, the output will be the event signatures $p^{(e)}$ and episode signatures $\theta^{(\mathcal{E})}$, where an episode signature is a probability distribution of the events in the episode. In other words, LDA assumes that each message in an episode is generated by first picking an event within an episode from the episode signature and then picking a message from the event based on the event signature. 

For our event mining problem, we are interested in event signatures and finding the start and finish times of each occurrence of an event. Therefore, the final step (which we describe next) is to extract the start and finish times from the episode signatures.

{\bf Putting it all together:} 
In order to detect all the episodes in which the event $e$ occurs prominently, we proceed as follows. We collect all episodes $\mathcal{E}$ for which the event occurrence probability $\theta^{(\mathcal{E})}_e$ is greater than a certain threshold $\eta>0$. We declare the start and finish times of the collected episodes as the start and finish times of the various occurrences of the event $e$. If an event spans many contiguous episodes, then the start time of the first episode and the end time of the last contiguous episode can be used as the start and finish time of this occurrence of the event. However, for simplicity, this straightforward step in not presented in the detailed description of the algorithm in \textsc{Algorithm~\ref{alg:CD-LDA}}.

   \begin{algorithm}
  	\begin{algorithmic}[1]
  		
  		\State \textbf{Input}: data points $\mathcal{D}$, threshold of occurrence of an event in an episode $\eta,$ the minimum value of TV distance $\delta$, minimum episode length $\alpha$.
  		
  		\State \textbf{Output}: Event signatures $p^{(1)},p^{(2)},\ldots,p^{(E)}$, Start and finish time $S_e,F_e$ for each event $e$.

  		\State Change points $\tau_1,\ldots, \tau_k \gets$ CD$(\mathcal{D},\alpha,\delta)$. Episode $\mathcal{E}_i \gets \{X_{\tau_{i-1}},\ldots, X_{\tau_{i}}\}$ for $i = 1$ to $k+1$.
  		
  		\State $p^{(1)},\ldots,p^{(E)} ; \theta^{(\mathcal{E}_1)},\ldots, \theta^{(\mathcal{E}_{k+1})} \gets $LDA$(\mathcal{E}_1,\ldots,\mathcal{E}_{k+1})$
  		
  		\State Consider event $e$. $\mathcal{G}_e \gets$  Set of all episodes $\mathcal{E}$ such that $\theta^{(\mathcal{E})}_e>\eta$. $S_e,F_e \gets$ start and finish times of all episodes in set $\mathcal{G}_e$. 
  	\end{algorithmic}
  	\caption{CD-LDA$(\mathcal{D},\alpha,\delta,\eta)$}
  	\label{alg:CD-LDA}
  \end{algorithm}

\begin{remark}
 There are many inference techniques for the LDA model, \cite{gibbs,LDA, vr1,vr2,anima,kannan}. We use the Gibbs sampling based inference from \cite{gibbs} on the LDA model. For a discussion on the comparison between the above methods, see Appendix \ref{app:ldamodel} in the supplementary material.
\end{remark}

\begin{remark}
	{\em CD-LDA algorithm works without knowledge of topology graph of message-generating elements. If topology graph is known, then  the algorithm can be improved as follows. We can run change-detection phase separately for messages restricted to each element and its graph neighbors (either single-hop or two-hop neighbors). The union of change-points could be used in the subsequent LDA phase. Since impact of an event is usually restricted to few hops within the topology, such an approach detects change points better by eliminating several messages far from event-source.}
\end{remark}

Note that the LDA algorithm requires an input for the number of events $E$. However, one can run LDA for different values of $E$ and choose the one with maximum likelihood \cite{LDA}. Hence $E$ need not be assumed to be an input to CD-LDA. One can also use the Hierarchical Dirichlet Process (HDP) algorithm \cite{hdp} which is an extension of LDA and figure out the number of topics from the data. In our experiments, we use the maximum likelihood approach to estimate the number of events. {This is exaplined in section \ref{sec:topics}}.


\subsection{Analysis of CD}
\label{sec:analysis}
As mentioned earlier, the novelty in the CD-LDA algorithm lies in the connection we make to topic modeling in document analysis. In this context, our key contribution is an efficient algorithm to divide the data set of messages into episodes (documents). Once this is done, the application of the LDA of episodes (documents), consisting of messages (words) generated by events (topics) is standard. Therefore, the correctness and efficiency of the CD part of the algorithm will determine the correctness and efficiency of CD-LDA as a whole. We focus on analyzing the CD part of the algorithm in this section. Due to space limitations, we only present the main results here, and the proofs can be found in the supplementary material.

Section \ref{sec:complexity} shows that the computational complexity of CD algorithm is linear in the number of data points. Section \ref{sec:consistency} contains the asymptotic analysis of the CD algorithm while section \ref{sec:sample} has the finite sample results.

\subsubsection{Computational complexity of CD}
\label{sec:complexity}
In this section we discuss the computational complexities of Algorithm \ref{alg:onechangept} and Algorithm \ref{alg:multiplechangept}. We will first discuss the computational complexity of detecting a change point in case of one change point. Algorithm \ref{alg:onechangept} requires us to compute $\arg\max_{l} \D (l)$ for $1 \le l \le n$. From the definition of $\D(l)$ in \eqref{eq:Dhat}, we only need to compute the empirical probability estimates $\widehat{p}_L(l)$, $\widehat{p}_R(l)$, and the empirical mean of the inter arrival time $\widehat{\E} S_L(l)$, $\widehat{\E} S_R(l)$ for every value of $l$ between $1$ to $n$. 

We focus on the computation of $\widehat{p}_L(l)$, $\widehat{p}_R(l)$. Consider any message $m$ in the distribution. For each $m,$ we can compute $\widehat{p}_{L,m}(l)$, $\widehat{p}_{R,m}(l)$ in $O(n)$ for every value of $l$ by using neighbouring values of $\widehat{p}_{L,m}(l-1)$, $\widehat{p}_{R,m}(l-1)$. 
\begin{align}
    \widehat{p}_{L,m}(l) &= \frac{(l-1)\widehat{p}_{L,m}(l - 1) + \ind{X_{l-1} =m} }{l},\nonumber
    \\
    \widehat{p}_{R,m}(l) &= \frac{(n - l + 1)\widehat{p}_{R,m}(l-1) - \ind{ X_{l-1} =m }}{n-l}
    \label{eq:computephat}
\end{align}
The computation of  $\widehat{\E} S_L(l),$ $\widehat{\E} S_R(l)$ for every value of $l$ from $1$ to $n$ is similar.

Performing the above computations for all $M$ messages, results in a computational complexity of $O(nM).$
In the case of $k$ change points, it is straightforward to see that we require $O(nMk)$ computations. In much of our discussion, we assume $M$ and $k$ are constants and therefore, we present the computational complexity results in terms of $n$ only.

\textbf{Related work:}
Algorithm \ref{alg:multiplechangept} executes the process of determining change points hierarchically. This idea was inspired by the work in \cite{nonparamchpt}. However, the metric $\D$ we use to detect change points is different from  that of \cite{nonparamchpt}.
The change in metric necessitates a new analysis of the consistency of the CD algorithm which we present in the next subsection. Further, for our metric, we are also able to derive sample complexity results which are presented in a later subsection.

\subsubsection{The consistency of change-point detection}
\label{sec:consistency}
In this section we discuss the consistency of the change-point detection algorithm, i.e., when the number of data points $n$ goes to infinity one can accurately detect the location of the change points.
In both this subsection and the next, we assume that the inter-arrival times of messages within each episode are i.i.d., and are independent (with possibly different distributions) across episodes.

\begin{theorem}
For $\tg \in (0,1),$ $D(\tg)=\lim_{n\rightarrow\infty}\D(\tg n)$ is well-defined and $D(\tg)$ attains its maximum at one of the change points if there is at least one change point.\label{lemma:consistency}
\end{theorem}
\begin{remark}
The proof  of the theorem \ref{lemma:consistency} for the single change-point case is relatively easy, but the proof in the case of multiple change points is rather involved. { So, due to space limitations, we only provide a proof of the single change point case and refer the interested reader to Appendix \ref{app:multiple} in the supplemental material for the proof of the multiple change point case.} 
\end{remark}
\begin{proof}
\textbf{Proof for single change point case:}
We first discuss the single change point case. Let the change point be at index $\tau$. The location of the change point is determined by the point where $\D(l)$ maximizes over $1<l<n$. We will show that when $n$ is large the argument where $\D(l)$ maximizes converges to the change point $\tau$.

Suppose all the points $X$ to the left of the change point $\tau$ are chosen i.i.d from a distribution $F$ and all the points from the right of $\tau$ are chosen from a distribution $G$, where $F\neq G$. Also, say the inter-arrival times $\Delta t_i$'s are chosen i.i.d from distribution $F_t$ and $G_t$ to the left and right of change point $\tau$, respectively. 
Let $l = \tg n$, $0<\tg <1$ be the index of any data point and $\tau = \g n$, the index of the change point. 
\\
\textbf{Case 1 $\tg\le\g$}: Suppose we consider the value of $\D(l) = \D(\g n)$ to the left of the actual change point, i.e, $l<\tau$ or $\tg<\g$. The distribution to the left of $\tg n$, $\widehat{p}_L(\tg n)$, has all the data points chosen from the distribution $F$. So $\widehat{p}_L(\tg n)$ is the empirical estimate for $F$. On the other hand, the data points to the right of  $\tg n$  come from a mixture of distribution $F$ and $G$. $\widehat{p}_R(\tg n)$ has $\frac{\g-\tg}{1-\tg}$ fraction of samples from $F$ and $\frac{1 - \g}{1-\tg}$ fraction of samples from $G$. Figure \ref{fig:TVdistance} below explains it pictorially.

\begin{figure}[ht]
\vspace{.3in}
\centerline{\includegraphics[scale = 0.35]{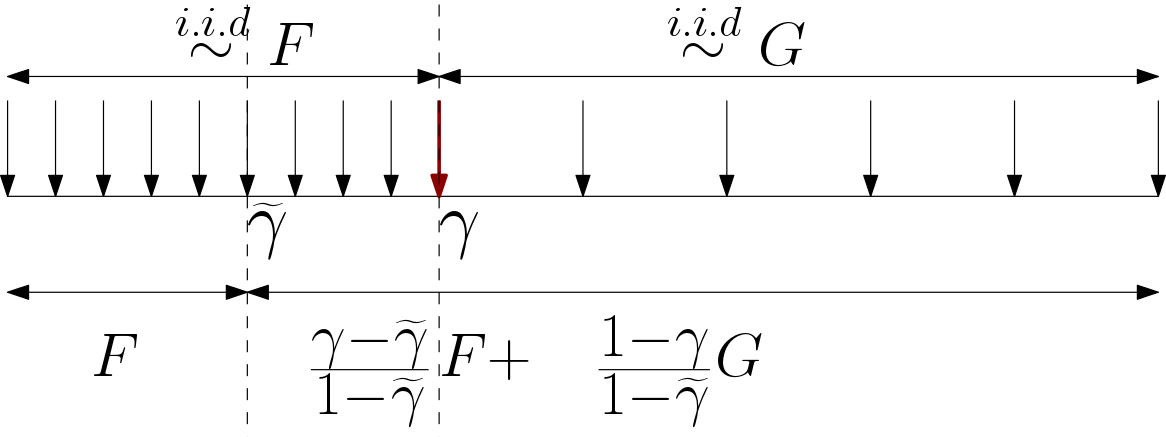}}
\vspace{.3in}
\caption{Consistency with two change points}
\label{fig:TVdistance}
\end{figure}

So $\widehat{p}_L(l)$ and $\widehat{p}_R$ defined in \eqref{eq:defp} converges to
\begin{align}
    \widehat{p}_L(l)\rightarrow F,\;\;\;& \widehat{p}_R(l)\rightarrow \frac{\g-\tg}{1-\tg}F + \frac{1 - \g}{1-\tg}G.\label{eq:pleft}
\end{align}

Similarly, we can say that the empirical mean estimates $\widehat{E}S_L(l)$ and $\widehat{E}S_R(l)$ converge to 
\begin{align}
    \widehat{E}S_L(l)\rightarrow \E F_t,\;\;\;& \widehat{\E}S_R(l)\rightarrow \frac{\g-\tg}{1-\tg}\E F_t + \frac{1 - \g}{1-\tg}\E G_t.\label{eq:Sleft}
\end{align}

We can combine \eqref{eq:pleft} and \eqref{eq:Sleft} to say that $\D(\tg n)\rightarrow D(\tg)$ where
\begin{align}
    \D(\tg n) &= \|\widehat{p}_L(\tg n) - \widehat{p}_R(\tg n)\| + |\E S_L(\tg n) - \E S_R(\tg n)|\nonumber\\
    &\rightarrow D(\tg) := \frac{1 -\g}{1-\tg}  (\| F - G\|_1 + |\E F_t - \E G_t|).
    \label{eq:Dleft}
\end{align}

Note that from the definition of $D$, $D(\g) = \| F - G\|_1 + |\E F_t - \E G_t|$.
\\
\textbf{Case 2 $\tg>\g$}:
Proceeding in a similar way to Case 1, we can show
\begin{align}
    \D(\tg n) &\rightarrow D(\tg) := \frac{\g}{\tg}  (\| F - G\|_1 + |\E F_t - \E G_t|).
    \label{eq:Dright}
\end{align}
\\
\\
From Case 1 and Case 2, we have
\begin{align}
    \tg\le\g, \;\;\;\;\;\; \D(\tg n )&\rightarrow D(\tg) = \frac{1 -\g}{1-\tg}D(\g )\nonumber 
    \\
    \tg>\g, \;\;\;\;\;\; \D(\tg n )&\rightarrow D(\tg ) = \frac{\g}{\tg}D(\g ). \label{Dlln}
\end{align}

Equation \eqref{Dlln} shows that the maximum of $D(\tg)$ is obtained at $\tg = \g$. 

\end{proof}

\subsubsection{The sample complexity of change-point detection}
\label{sec:sample}

In the previous subsection, we studied the CD algorithm in the limit as $n\rightarrow\infty.$ In this section, we analyze the algorithm when there are only a finite number of samples. For this purpose, we assume that the inter-arrival distribution of messages have sub-Gaussian tails.

We say that Algorithm CD is correct if the following conditions are satisfied. Let $\epsilon>0$ be a desired accuracy in estimation of the change point.

\begin{definition}
Given $\epsilon>0$, Algorithm CD is correct if
\begin{itemize}
    \item there are change points $0 <\frac{\tau_1}{n} = \g_1,\ldots, \frac{\tau_k}{n} = \g_k <1$ and the algorithm gives $\hg_1,\ldots,\hg_k$ such that $\max_{i}|\widehat{\g}_i - \g_i|<\epsilon$.
    \item there is no change point and $\D(\g n)<\delta,\forall\g\in\{\g_1,\ldots,\g_k\}$.
\end{itemize}
\label{def:correctmultiple}
\end{definition}

Now we can state the correctness theorem for Algorithm \ref{alg:multiplechangept}. The sample complexity is shown to scale logarithmically with the number of change points.

\begin{theorem}
Algorithm \ref{alg:multiplechangept} is correct in the sense of Definition \ref{def:correctmultiple} with probability $(1 - \beta)$ if
\[n  = \Omega\left(\max\left( \frac{\log\left(\frac{2k+1}{\beta}\right)}{\epsilon^2}, \frac{M^{1+c}}{\epsilon^{2(1+c)}}\right)\right), \]
for sufficiently small $\alpha,$ $\delta,$ $\epsilon$  and for any $c>0$.
\label{th:multiplechptSmaple}
\end{theorem}
\begin{remark}
The proof of this theorem uses the method of types and Pinsker's inequality. We present here the proof for the single change point case. Due to space constraints, we move the proof for multiple change points to Appendix \ref{sec:multChPt} in the supplementary material. 
\end{remark}

\begin{proof}
We first characterize the single change point case in finite sample setting. 
In order to get the sample complexity, we prove the correctness for Algorithm \ref{alg:onechangept} as per Definition \ref{def:correctmultiple} with high probability.
Before we go into the proof, we state the assumptions on $\alpha,\delta,\epsilon$ under which the proof is valid.

\begin{itemize}
    \item Suppose a change point exists at index $\g n$ and the metric $\D(\g n)$ converges to $D(\g)$ at the change point. Then $\epsilon$ can only be chosen in following region: $\epsilon$ has to be less than the value of the metric at the change point, $\epsilon<D(\g)$; $\epsilon$ has to be less than the minimum episode length, $\epsilon < \min(\g,1-\g)$. 
    \item If a change point exists at index $\g n$, $\alpha$ has to chosen less than the minimum episode length minus $\epsilon$, $\alpha < \min(\g,1-\g) - \epsilon$.
    \item The threshold $\delta < D(\g) - \epsilon$.
\end{itemize}

Under the above assumptions we show that Algorithm \ref{alg:onechangept} is correct as per the Definition \ref{def:correctmultiple} with probability at least  
$$1 - (6n+4)\exp\left(-\frac{\min (\delta,1)^2\epsilon^2\alpha^2}{512\max(\sigma^2,1)}n + M\log(n)\right).$$

Suppose $$\hg n = \arg \max_{\tg n} \D(\tg n) .$$
The idea is to upper bound the probability when Algorithm \ref{alg:onechangept} is not correct. From Definition \ref{def:correctmultiple} this happens when,

\begin{itemize}
    \item Given a change point exists at $\g\in (0,1)$, $$(\D(\hg n )>\delta,|\g - \widehat{\g}|<\epsilon, \alpha<\hg<1 - \alpha)^c$$ occurs. Say the event $E_1$ denotes $E_1 = \{\D(\widehat{\g})>\delta,|\g - \widehat{\g}|<\epsilon, \alpha<\hg<1 - \alpha\}$.
    \item Given a change point does not exist, $$\D(\widehat{\g})>\delta, \alpha<\hg<1 - \alpha.$$ When a change point does not exist we write $\g = 0$. Say the event $E_2$ denotes $E_2 = \{\g =0,\alpha<\hg<1 - \alpha\}$
\end{itemize}

So
\begin{align}
    &P(\text{Algorithm \ref{alg:onechangept} is NOT correct})\nonumber
    \\
    &\le P(E_1^c|0<\g<1) + P(\D(\hg)>\delta|E_2 )\label{eq:algcorrect}.
\end{align}

We analyze each part in \eqref{eq:algcorrect} separately. 

\textbf{Case 1:} Suppose no change point exists and say all the data points are drawn from the same  mutinomial distribution $F$ and all inter-arrival times are generated i.i.d from a distribution $F_t$. 
Given event $E_2$, if $\| \widehat{p}_L(\hg n) - F \|, \| \widehat{p}_R(\hg n) - F \|, |\widehat{\E} S_L(\hg n) - \E F_t |, |\widehat{\E} S_R(\hg n) - \E F_t|$ are all less than $\delta/4$, then $\D(\hg)<\delta$. So $P(\D(\hg)>\delta|E_2 ) \le P(\| \widehat{p}_L(\hg n) - F \|>\delta/4 | E_2) + P(\| \widehat{p}_R(\hg n) - F \|>\delta/4 | E_2) + P(|\widehat{\E} S_L(\hg n) - \E F_t | > \delta/4|E_2) + P(|\widehat{\E} S_R(\hg n) - \E F_t|>\delta/4|E_2)$. Now,
we can use Sanov's theorem followed by Pinsker's inequality to upper bound each of the above terms as  
\begin{align}
    &P(\D(\hg)>\delta|E_2 )\le (n\hg + 1)^M\exp( - n\delta^2/16)\nonumber\\
    &+ ((1 - \hg)n + 1)^M\exp( - n\delta^2/16)+ 2\exp( - \alpha n\delta^2/32\sigma^2)\nonumber
    \\
    &  + 2\exp( - \alpha n\delta^2/32\sigma^2)\nonumber
    \\
    &\le 4(n+2)^M\exp\left(-n \frac{\alpha\delta^2}{32\max(\sigma^2,1)}\right).\label{eq:nochpt}
\end{align}

\textbf{Case 2:}
Next, we look at the case when a change point exists at $\g n$. Say the messages are drawn from a distribution $F$ to the left of the change point and $G$ to the right of the change point. Also, suppose the inter-arrival time distribution to the left of the change point is $F_t$ and the inter-arrival time distribution to the right is $G_t$. According to our assumptions,  $\alpha$ is chosen such that $\alpha + \epsilon<\g<1 - (\alpha +\epsilon)$. Hence 
\begin{align}
    &P(E_1^c|0<\g<1)\le P(\D(\hg n )<\delta|0<\g<1)\nonumber
    \\
    & + P(|\hg - \g|>\epsilon | \D(\g)>\delta,0<\g<1)\nonumber\\
    & + P(\alpha<\hg<1 - \alpha | \D(\g)>\delta, |\hg - \g|<\epsilon,0<\g<1 ).\label{eq:chpt}
\end{align}
Given the assumption on $\alpha$, $P(\alpha<\hg<1 - \alpha | \D(\g)>\delta, |\hg - \g|<\epsilon,0<\g<1 ) = 0$. 
The rest of the proof deals with upper bounding $P(\D(\hg n )<\delta|0<\g<1)$ and $P(|\hg - \g|>\epsilon | \D(\g)>\delta,0<\g<1)$.

In lemma \ref{lemma:conc1}-\ref{lemma:conc3} we develop the characteristics of $\hg$ and $D(\hg)$ when a change point exists at $\g n$. Lemma \ref{lemma:conc1}-\ref{lemma:conc3} are proved in Appendix \ref{app:l2},\ref{app:l3l4} of the supplementary material. First, we analyze the concentration of $\D(\tg n)$ for any value of $\tg$ in the Lemma \ref{lemma:conc1}.

\begin{lemma}
$|\D(\tg n) - D(\tg  )|\le \epsilon$ w.p. at least $ 1 - 3n\exp\left(  -\frac{\epsilon^2\alpha^2}{128\sigma^2}n + M\log(n)   \right)$ for all values of $\tg$ when $\D(\tg n)$ is defined.
\label{lemma:conc1}
\end{lemma}
Lemma \ref{lemma:conc1} shows that the empirical estimate $\D(\tg n)$ is very close to the asymptotic value $D(\tg )$ with high probability. Recall that the argument at which $\D$ maximizes is $\hg n$. we next show in Lemma \ref{lemma:conc3} that the value of  metric $D$ at $\hg$ is very close to the value of the $D$ at the change point $\g$.
\begin{lemma}
$|D(\gamma) - D(\hg)|<2\epsilon$ w.p. $ 1 - 3n\exp\left(  -\frac{\epsilon^2\alpha^2}{128\sigma^2}n + M\log(n)   \right)$\label{lemma:conc2}
\end{lemma}
Finally, in Lemma \ref{lemma:conc3} we show that $\hg$ is close to the change point $\g$ with high probability.
\begin{lemma}
$ |\hg - \g|<\epsilon $ w.p. $1 - 3n\exp\left(  -\frac{\epsilon^2D^2(\g)\alpha^2}{512\sigma^2}n + M\log(n)   \right)$.\label{lemma:conc3}
\end{lemma}

Also, using lemma \ref{lemma:conc2} and assuming that $\delta$ is chosen such that $\delta < D(\g)-\epsilon$,
\begin{align}
    &P(\D(\hg n )<\delta|0<\g<1)\nonumber
    \\
    &\le P(\D(\hg n )<\delta|0<\g<1,|\D(\hg n) - D(\g)|<\epsilon)\nonumber\\
    &+ P(|\D(\hg n) - D(\g)|>\epsilon)\nonumber\\
    &\le 0 + 3n\exp\left(  -\frac{\epsilon^2\alpha^2}{128\sigma^2}n + M\log(n)   \right)\label{eq:chpt1}
\end{align}
Lemma \ref{lemma:conc3} gives a bound on $P(|\hg - \g|>\epsilon | \D(\g)>\delta,0<\g<1)$. Using this along with \eqref{eq:chpt1} in \eqref{eq:chpt} we have
\begin{align}
    &P(E_1^c|0<\g<1)\le 3n\exp\left(  -\frac{\epsilon^2\alpha^2}{128\sigma^2}n+ M\log(n)   \right)\nonumber\\
    & + 3n\exp\left(  -\frac{\epsilon^2D^2(\g )\alpha^2}{512\sigma^2}n + M\log(n)  \right).\label{nn}
\end{align}

Finally, putting together  \eqref{eq:nochpt} and \eqref{nn} into \eqref{eq:algcorrect}, we have
\begin{align}
    &P(\text{Algorithm \ref{alg:onechangept} is NOT correct})\nonumber\\
    &\le (6n+4)\exp(-\frac{\min(\delta,1)^2\epsilon^2\alpha^2}{512\max(\sigma^2,1)} + M\log(n+2))\label{eq:sample1}
\end{align}
Ignoring the constants in \eqref{eq:sample1}, we can derive the sample complexity result for the one change point case.
\end{proof}

\section{ Evaluation with Real Datasets}
\label{sec:exp}
We now present our experimental results with real data sets from large operational network. The purpose of  experiments is three-fold. First, we wish to compare our proposed CD-LDA algorithm with other techniques proposed (adapted to our setting) in the literature. Second, we want to validate our results against manual expert-derived event signature for a prominent event. Third, we want to understand the scalability of our method with respect to very large data sets.

{\bf Datasets used:} We use two data sets: one from a legacy network of physical elements like routers, switches etc., and another from a recently deployed virtual network function (VNF). The VNF dataset is used to validate our algorithm by comparing with expert knowledge. The other one is used to show that our algorithm is scalable, i.e., it can handle large data sets and it is less sensitive to the hyper parameters.

\begin{itemize}[leftmargin=15pt]
 \item {\bf Dataset-1:} This data set consists of around 97~million raw syslog messages collected from~3500 distinct physical network elements (mostly routers) from a nationwide operational network over a 15-day period in~2017. There are {$39330$} types of messages.
 \item {\bf Dataset-2:} The second data set  consists of  around $728,000$ messages collected from $285$ distinct physical/virtual network elements over a 3~month period from a newly deployed {\em virtual network function} (VNF) which is implemented on a data-center using multiple VMs. 
\end{itemize}
We implemented the machine-learning pipeline as shown in Figure~\ref{fig:mlpipeline}. The main algorithmic component in the figure shows CD-LDA algrothm; however, for the purpose of comparison, we also implemented two additional algorithms described shortly. Before the data is applied to any of the algorithms, there are two-steps, namely, Template-extraction (in case of textual syslog data) and pre-processing (for both syslog and alarms). These steps are described in Appendix \ref{app:preprocessing} in the supplementary material.

\subsection{Benchmark Algorithms}
We compare CD-LDA with the following algorithms.

\subsubsection{Algorithm B: A Bayesian inference based algorithm}
We consider a fully Bayesian inference algorithm to solve the problem. A Bayesian inference algorithm requires some assumptions on the statistical generative model by which the messages are generated. Our model here is inspired by topic modelling across documents generated over multiple eras\cite{tot}. Suppose that there are $E$ events which generated our data set, and event $e$ has a signature $p^{(e)}$ as mentioned earlier.  The generative model for generating each message is assumed to be as follows. 
\begin{itemize}[leftmargin=15pt]
	\item To generate a message, we first assume that an event $e \in[1,2,\ldots, E]$ is chosen with probability $P_e.$ 
	\item Next, a message $m$ is chosen with probability $p_m^{(e)}.$ 
	\item Finally, a timestamp is associated with the message which is chosen according to a beta distribution $\beta(a_e,b_e),$ where the parameters of the beta distribution are distinct for different events.
\end{itemize}
The parameters of the generative model ${P_e, p^{(e)}_m, a_e, b_e}$ are unknown. As in standard in such models, we assume a prior on some of these parameters. Here, as in \cite{tot}, we assume that there is a prior distribution on $q$ over the space of all possible $P$ and a prior $r$ over the space of all possible $p^{(e)}.$ The prior $r$ is assumed to be independent of $e.$ Given these priors, the Bayesian inference problem becomes a maximum likelihood estimation problem, i.e.,
$$\max_{{a_e, b_e, p^{(e)}}_e, P} \mathbb{P}_{q,r} (\mathcal{D} | P, \{p^{(e)}\}_e).$$
We use Gibbs sampling to solve the above maximization problem.
There are two key differences between Algorithm~B and proposed CD-LDA. CD-LDA first breaks up the datasets into smaller episodes whereas Algrothm-B uses prior distributions (the beta distributions) to model the fact that different events happen at different times. We show that, such an algorithm works, but the inference procedure is dramatically slow due to additional parameters to infer $\{a_e,b_e\}_e$. 

\subsubsection{Algorithm C: A Graph-clustering based algorithm}
\label{sec:algC}
For the purposes of comparison, we will also consider a very simple graph-based clustering-based algorithm to identify events. This algorithm is inspired from graph based clustering used in event log data in \cite{algC}. The basic idea behind the algorithm is as follows: we construct a graph whose nodes are the messages in the set $\mathcal{M}.$  We divide the continuous time interval $[0,T]$ into $T/w$ timeslots, where each timeslot is of duration $w.$ For simplicity, we will assume that $T$ is divisible by $w.$ We draw an edge between a pair of nodes (messages) and label the edge by a distance metric between the messages, which roughly indicates the likelihood with which two messages are generated by the same event. Then, any standard distance-based clustering algorithm on the graphs will cluster the messages into clusters, and one can interpret each cluster as an event. Clearly, the algorithm has the following major limitation:
it can detect $\mathcal{M}_e$ for an event $e$ and not $p^{(e)}.$ In some applications, this may be sufficient. Therefore, we consider this simple algorithm as a possible candidate algorithm for our real data set.

We now describe how the similarity metric is computed for two nodes $i$ and $j.$ Let $N_i$ be the number of timeslots during which a message $i$ occurs and let $N_{ij}$ be the number of timeslots during which both $i$ and $j$ appear in the same timeslot. Then, the distance metric between nodes $i$ and $j$ is defined as
$$\rho_{ij}=1-\frac{N_{ij}}{N_i+N_j-N_{ij}}.$$
Thus, a smaller $\rho_{ij}$ indicates that $i$ and $j$ co-occur frequently. The idea behind choosing this metric is as follows: messages generated by the same event are likely to occur closer together in time. Thus, $\rho_{ij}$ being small indicates that the messages are more likely to have been generated by the same event, and thus are closer together in distance. 
\subsection{Results: Comparison with Benchmark Algorithms}
\label{sec:visualValidation}
For the purposes of this section only, we consider a smaller slice of data from Dataset-1. Instead of considering all the 97~million messages, we take a small slice of 10,000~messages over a 3~hour duration from 135~distinct routers. Let us call this data set $\mathcal{D}_s$. There are two reasons for considering this smaller slice. 
Firstly, it is easier to visually observe the ground truth in this small data set and verify visually if CD-LDA is giving us the ground truth. We can also compare the results from different methods with this smaller data set.
Secondly, as we show later in this section, the Bayesian inference Algorithm-B is dramatically slow and so running it over the full dataset is not feasible. Nevertheless, the smaller dataset allows us to validate the key premise behind our main algorithm, i.e., the decomposition of the algorithm into the CD and LDA parts.
\begin{figure}[t]
    \centering
	\begin{subfigure}{0.4\textwidth}
		
		\includegraphics[width=1\linewidth]{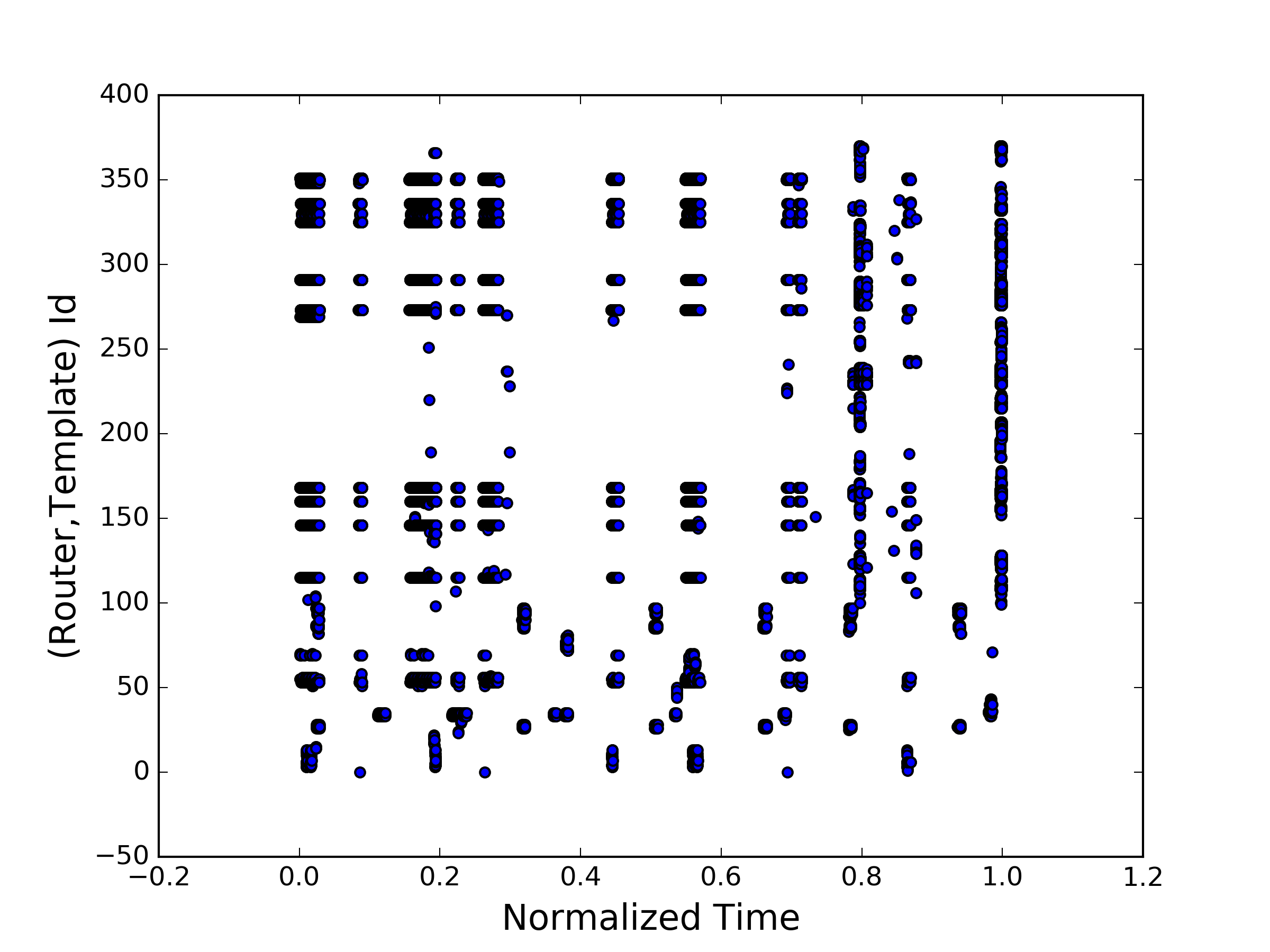}
		\caption{}
		\label{fig:dataviz}
	\end{subfigure}%
	
	\begin{subfigure}{0.4\textwidth}
		\centering
		\includegraphics[width=1\linewidth]{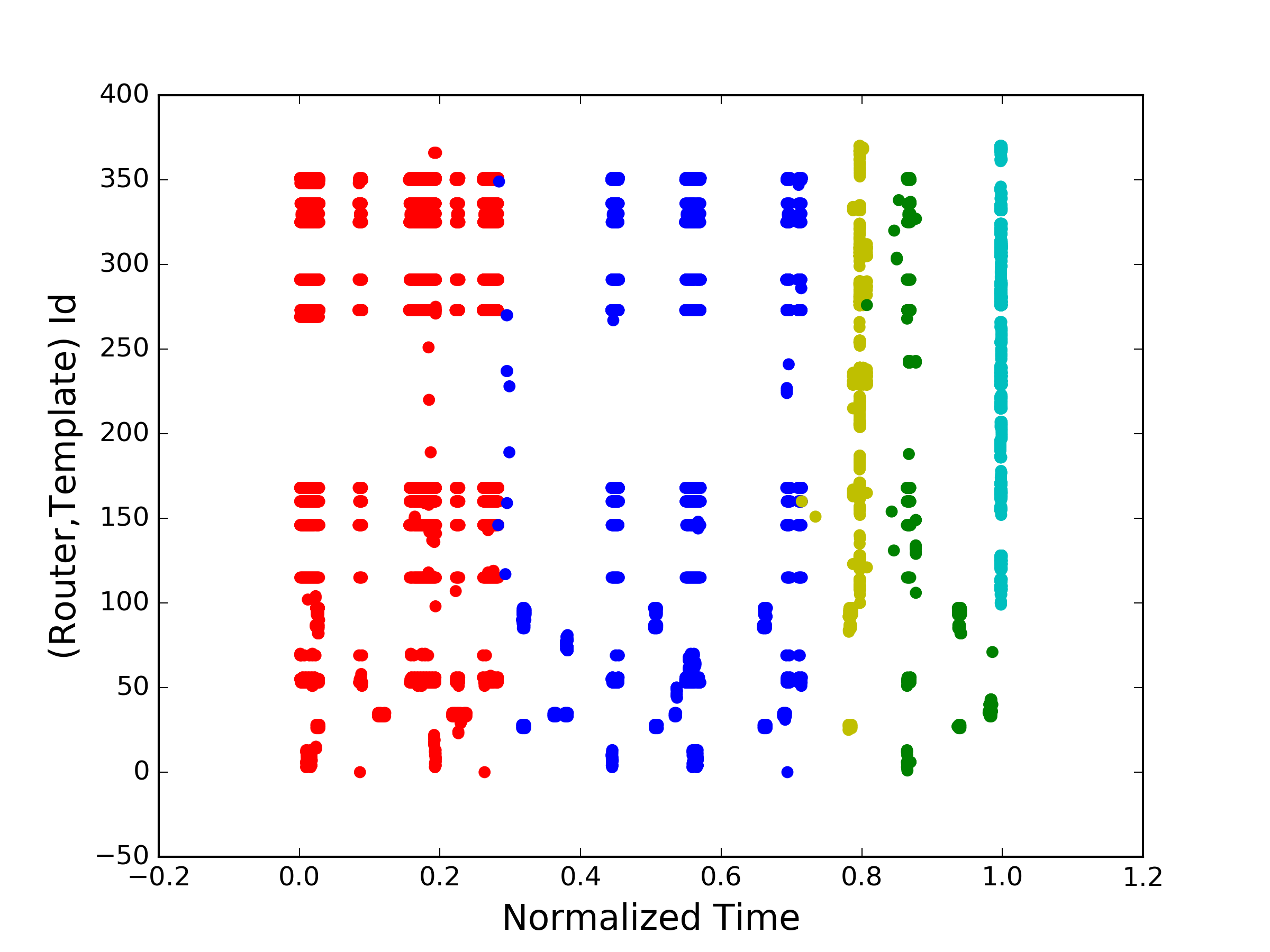}
		\caption{}
		\label{fig:episodes}
		
	\end{subfigure}
	\caption{Top panel shows scatter plot of different message-ids over the period of comparison and bottom panel shows the episodes detected by CD phase of Algorithm~CD-LDA.}
 	\label{fig:scatter}
 	\vspace{-0.2in}
\end{figure}

{\bf Applying CD-LDA on this dataset slice:} Figure~\ref{fig:scatter}a shows the data points in x-axis and the message-ids on y-axis.  Figure~\ref{fig:scatter}b shows the~$5$ episodes after the CD part of CD-LDA, where we chose $\alpha=0.1$ and $\delta=0.5.$ For the LDA part, instead of specifying the number of events, we use maximum likelihood to find the optimal number of events and based on this, the number of events was found to be $2.$

We next compare event signatures produced by CD-LDA with Algorithm~B and Algorithm~C.

{\bf CD-LDA versus Algorithm B:}
For all unknown distributions, we assume a uniform prior in Algorithm B. Algorithm B is run with input number of events as $2,3,4,5$. It  turns out that, with $3$ events the algorithm converges to a solution which has maximum likelihood. However, upon clustering the event signatures $p^{(e)}$ based on TV-distance between the event signatures, we find only two events. 
\textit{The maximum TV-distance between the events signatures found from the two algorithms is $0.068$}. Hence, we can conclude that the event signatures found by both the algorithms are very similar.

\begin{figure}[ht]
	\vspace{-0.1in}
	\centerline{\includegraphics[scale = 0.33]{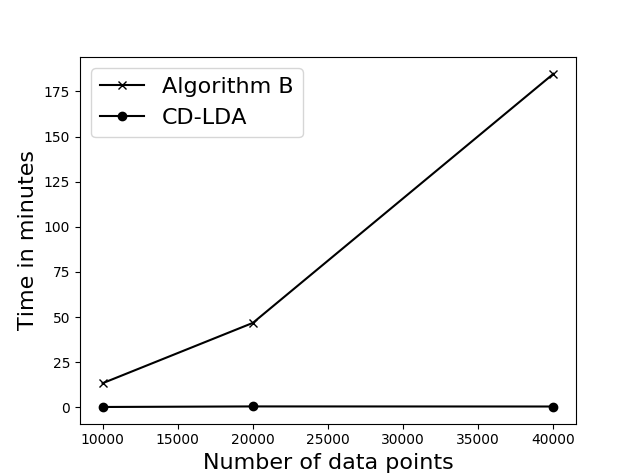}}
	\caption{Time performance: CD-LDA vs Algrithm B}
	\label{fig:CDLDAvsBtime}
\end{figure}

Despite the fact that Algorithm B using fewer hyper-parameters, it is not fast enough to run on large data sets. Figure \ref{fig:CDLDAvsBtime}a shows the time taken by CD-LDA and Algorithm B as we increase the size of the data set from $10,000$ to $40,000$ points. With $40,000$ data points and $12$ events as input Algorithm B takes 3 hours whereas CD-LDA only takes 26.57 seconds. Clearly, we cannot practically run Algorithm B on large data sets with millions of points. 

{\bf CD-LDA versus Algorithm C:}
In this section we compare CD-LDA versus algorithm C on data set $\mathcal{D}_s$. Algorithm C can produce the major event clusters as CD-LDA, but does not provide the start and end time for the events.
We form the co-occurrence graph for Algorithm C with edge weight as described in section \ref{sec:algC} and nodes as messages which occur more than at least $5$ times in the data set $\mathcal{D}_s$. All the edges with weight more than $0.6$ are discarded and we run a clique detection algorithm in the resulting graph. 

We quantitatively compare the event signature $\mathcal{M}_e$ of the top two cliques found by Algorithm C with those found by CD-LDA. 
Suppose that message sets identified by Algorithm C for the two events are $\mathcal{M}_{e1}$ and $\mathcal{M}_{e2}$ respectively. Message sets (messages with probability more than $0.007$) identified by CD-LDA for the two events are denoted by $\mathcal{S}_{e1}$ and $\mathcal{S}_{e2}$. We can now compute the Jaccard Index between the two sets.
\begin{center}
\begin{tabular}{l l}
     $\dfrac{|\mathcal{M}_{e1}\cap \mathcal{S}_{e1}|}{|\mathcal{M}_{e1}\cup \mathcal{S}_{e1}|} = 0.73$ & $\dfrac{|\mathcal{M}_{e2}\cap \mathcal{S}_{e2}|}{|\mathcal{M}_{e2}\cup \mathcal{S}_{e2}|} = 0.68.$  \\
\end{tabular}
\end{center}
Since the full Bayesian inference (Algorithm B) agrees with CD-LDA closely, we can conclude that Algorithm C gets a large fraction of the messages associated with the event correctly. However, it also misses a significant fraction of the messages, and additionally Algorithm C does not provide any information about start and end times of the events. Also, the events found are sensitive to the threshold for choosing the graph edges, something we have carefully chosen for this small data set.

\subsection{Results: Comparison with Expert Knowledge and Scalability}
\label{sec:vnf}

\begin{table}[t]
	\begin{center}
	\scriptsize
		\begin{tabular}{  l l l}
			\normalsize{Event $1$} & \normalsize{Event $2$}\;\;\;\;\;\;\;\;\;\;\;\;\ldots & \normalsize{Event $8$}\\
			\texttt{mmscRuntimeError} & \texttt{ISCSI\_multipath} & \texttt{SNMP\_sshd} \\
			\texttt{SUDBConnectionDown} & \texttt{Logmon\_contrail} & \texttt{SNMP\_crond}\\
			\texttt{SocketConnectionDown} & \texttt{VRouter-Vrouter} & \texttt{SNMP\_AgentCheck}\\
			\texttt{SUDBConnectionUp} & \texttt{LogFile\_nova} & \texttt{SNMP\_ntpd}\\
			\texttt{SocketConnectionUp} & \texttt{SUDBConnectionDown} & \texttt{SNMP\_CPU}\\
			\texttt{mmscEAIFUnavailable} & \texttt{IPMI} & \texttt{SNMP\_Swap}\\
			\texttt{bigipServiceUp} & \texttt{bigipServiceDown} & \texttt{SNMP\_Mem}\\
			\texttt{bigipServiceDown} & \texttt{bigipServiceUp} & \texttt{SNMP\_Filespace}\\
			\texttt{SNMP\_Mem} & \texttt{HW\_IPMI} & \texttt{Ping\_vm}
		\end{tabular}
	\end{center}
	\caption{Events generated by CD-LDA and the constituent messages in decreasing order of probability. Event 8 matches with expert provided event signature.}
	\label{tab:ldavnf}
\end{table}

{\bf Validation by comparing with manual event signature:} The intended use-case of our methodology is for learning events where the scale of data and system does not allow for manual identification of event signatures. However, we still wanted to validate our output against a handful of event signatures inferred manually by domain experts. For the purpose of this section, we ran CD-LDA for Dataset-2 which is for an operational VNF. For this data set, an expert had identified that a known service issue had occurred on two dates: 11-Oct and 26-Nov, 2017. This event generated messages with Ids 
\texttt{Ping\_vm}, 
\texttt{SNMP\_AgentCheck}, \texttt{SNMP\_ntpd}, \texttt{SNMP\_sshd}, \texttt{SNMP\_crond}, \texttt{SNMP\_Swap}, \texttt{SNMP\_CPU}, \texttt{SNMP\_Mem},
\texttt{SNMP\_Filespace}.

We ran CD-LDA on this data set with parameters $\alpha=0.01$ and $\delta=0.1$. We chose $10$~events for the LDA phase {by looking at the likelihood computed using cross validation for different number of topics. See section \ref{sec:topics} for details of the maximium likelihood approach}. 
Table~\ref{tab:ldavnf} shows the events detected by CD-LDA in decreasing order of probability. Also, top $9$ messages are listed for each event. Indeed, we note that {\em Event~$8$ resembles the expert provided event. CD-LDA detected this event as having occurred from 2017-10-08 17:35 to 2017-10-17 15:55 and 2017-11-25 13:45 to 2017-11-26 03:10. } The longer than usual detection window  for 11-Oct is due to the fact that there were other events occurring simultaneously in the network and the Event $8$ contributed to small fraction of messages generated during this time window.
Finally, as shown in Table~\ref{tab:ldavnf}, our method also discovered several event signatures not previously known.

{\bf Scalability and sensitivity:} To understand the scalability of CD-LDA with data size, we ran it on Dataset-1 with about $97$ million
 data points. CD-LDA was run with the following input: $\alpha=0.01$, $\delta= 0.1,$ and  the number of events equal to $20.$ The CD part of the algorithm detects $57$ change points. The sensitivity of this output with respect to $\alpha,$ $\delta$ is discussed next. The event signatures are quite robust to these parameter choice, but as expected, the accuracy of the start and finish time estimates of the events will be poorer for large values of $\alpha$ and $\delta.$ Overall, CD-LDA takes about $6$ hours to run, which is quite reasonable for a dataset of this size. Reducing the running time by using other methods for implementing LDA, such as variational inference, is a topic for future work.

\begin{figure}
    \centering
    \includegraphics[scale = 0.35]{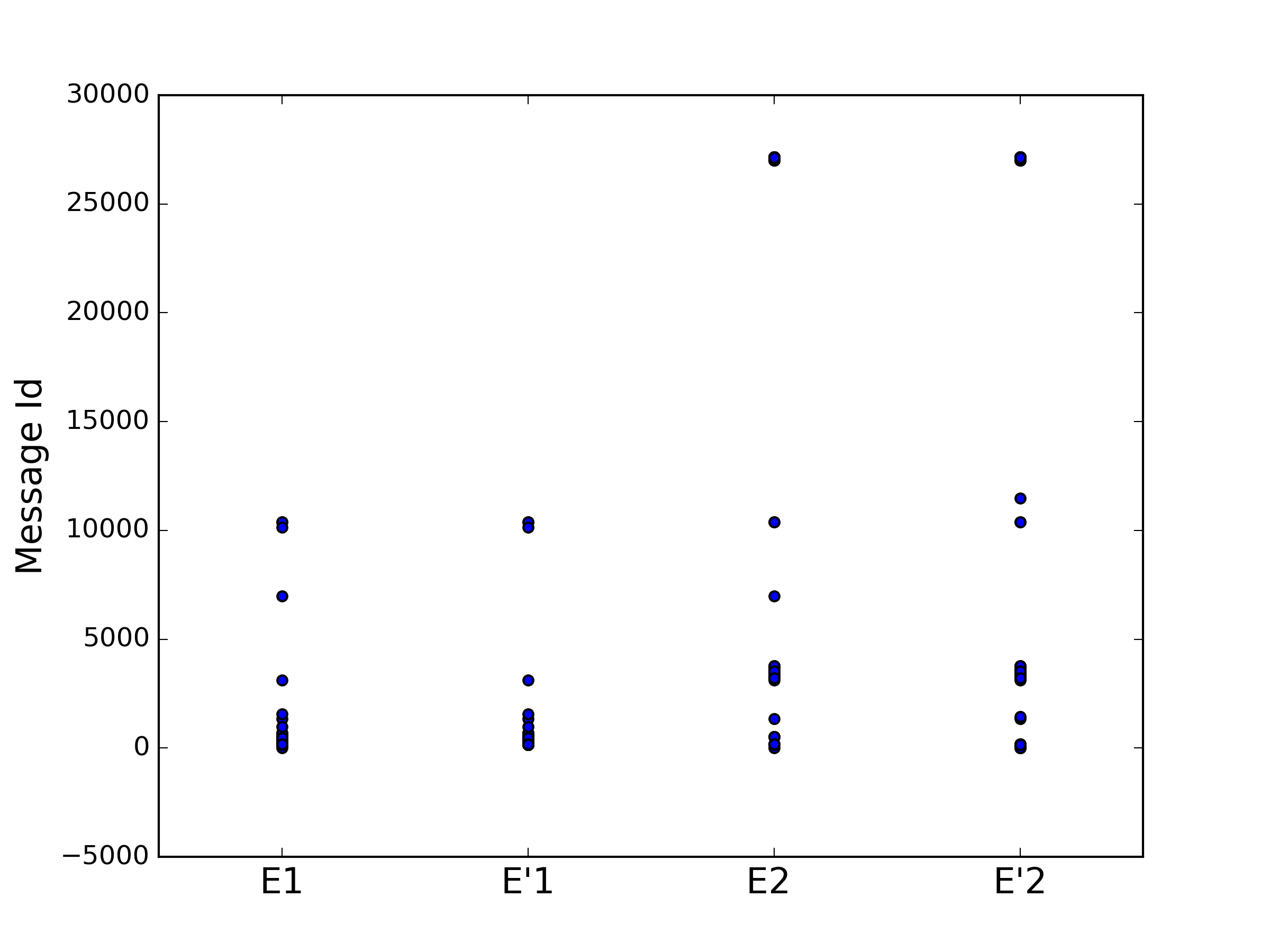}
    \caption{Comparison of Event signatures for first two events with $\alpha_1,\delta_1$(E1,E2) vs $\alpha_2,\delta_2$(E'1,E'2). }
    \label{fig:syslog}
\end{figure}

Parameter $\alpha$ specifies the minimum duration of episode that can be detected in the change detection. By increasing $\delta$ we can control to detect the more sharp change points (change points across which the change in distribution is large), and decreasing $\delta$ helps us detect the soft change points as well. So $\alpha,\delta$ control the granularity of the change point detection algorithm. Parameter $\eta$ is a user defined parameter to detect the episodes in which a particular event occurs. We demonstrate the sensitivity of CDLDA to $\alpha$ and $\delta$. We run CDLDA with $\alpha_2 = 10\%, \delta_2 = 0.5$ on Dataset-2 and compare it with results when run with parameters $\alpha_1 = 1\%, \delta_1 = 0.1$. Table \ref{tab:syslog} shows the first two events for parameters $\alpha_1,\delta_1$ when compared to first two events for parameter $\alpha_2,\delta_2$. CDLDA detects 57 change points with $\alpha_1,\delta_1$ whereas it only detects $19$ change points with $\alpha_2, \delta_2$. Despite this, Figure \ref{fig:syslog} and {Table \ref{tab:quant}} shows that the event signatures for the first two events are almost the same. But, since the episodes are larger in duration with $\alpha_2,\delta_2$, the start and end times of the first two events are less accurate than $\alpha_1,\delta_1$. In particular, event 2 is shown to occur from 2-10 05:00 to 2-14 00:00 with  $\alpha_2,\delta_2$ in Table \ref{tab:syslog} whereas it broken into two episodes, 2-10 05:00 to 2-10 13:33 and 2-10 15:27 to 2-14 00:00 , with $\alpha_1,\delta_1$.

\begin{table}
\begin{center}
\begin{tabular}{|c|c|c|c|}                     
    \hline
    \multicolumn{4}{|c|}{$\alpha_1 = 1\%,\delta_1 = 0.1$ vs $\alpha_2 = 10\%,\delta_2 = 0.5$} \\
    \hline
                                             $\frac{|\mathcal{M}_1 \Delta \mathcal{M}'_1|}{|\mathcal{M}_1 \cup \mathcal{M}'_1|}$ &  $\frac{|\mathcal{M}_2 \Delta \mathcal{M}'_2|}{|\mathcal{M}_2 \cup \mathcal{M}'_2|}$& TV dist in $p^{(1)}$ & TV dist in $p^{(2)}$ \\
    \hline
                                                0.046                               &0.077 & 0.036 &0.08\\
    \hline
\end{tabular}
\end{center}
\caption{Comparing results of CD-LDA for different values of $\alpha,\delta$}
\label{tab:quant}
\begin{center}
\scriptsize
    \begin{tabular}{c c}
        \normalsize{Event $1$} & \normalsize{Event $2$}\\
         2017-02-14 00:00 to 2017-02-15 23:59 & 2017-02-06 19:29 to 2017-02-07 16:42 \\
         & 2017-02-08 00:00 to 2017-02-08 06:25\\
         & 2017-02-08 23:59 to 2017-02-10 04:07\\
         & \underline{2017-02-10 05:00 to 2017-02-14 00:00}\\
    \end{tabular}
\end{center}
\caption{Results of CD-LDA on Dataset-2 with $\alpha_2=10\%,\delta_2 = 0.5$ }
\begin{center}
\scriptsize
    \begin{tabular}{c c}
        \normalsize{Event $1$} & \normalsize{Event $2$}\\
         2017-02-14 00:00 to 2017-02-15 23:59 & 2017-02-05 06:21 to 2017-02-07 16:42 \\
         & 2017-02-08 00:00 to 2017-02-10 00:00\\
         & 2017-02-10 03:07 to 2017-02-10 04:07\\
         & \underline{2017-02-10 05:00 to 2017-02-10 13:33}\\
         & \underline{2017-02-10 15:27 to 2017-02-14 00:00}
    \end{tabular}
\end{center}
\caption{Results of CD-LDA on Dataset-2 with $\alpha_1=1\%,\delta_1 = 0.1$}
\label{tab:syslog}
\end{table}

\subsubsection{Selection of the number of topics in LDA}
\label{sec:topics}
\begin{figure}
    \centering
    \includegraphics[scale = 0.35]{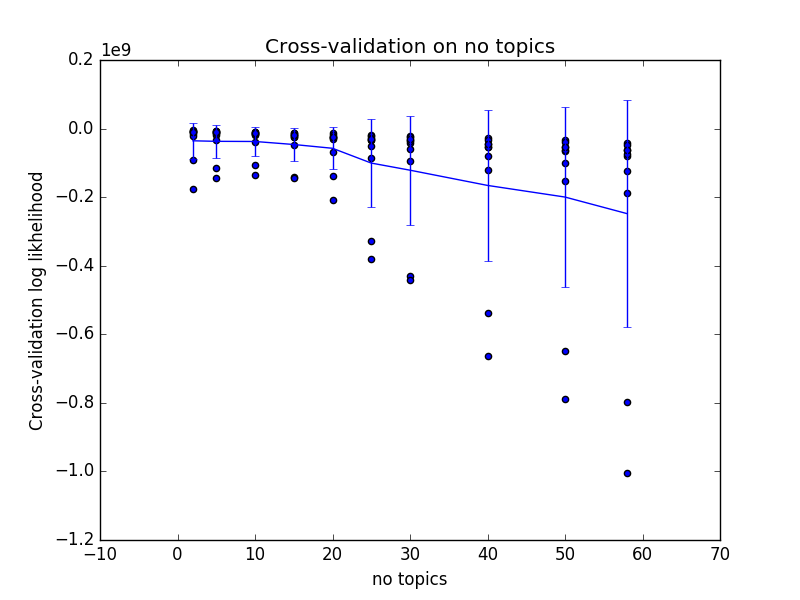}
    \caption{Likelihood vs number of topics in Dataset-1}
    \label{fig:cross}
\end{figure}

{
For Dataset-1, we do 10-fold cross validation. We group the 58 documents found by change detection into 10 sets randomly. We compute the likelihood on one group with a model trained using documents in the remaining 9 groups. We plot the average likelihood in Figure \ref{fig:cross} vs the number of topics. There is a decrease in likelihood around $20$ and hence, we choose the number of topics as $20$.}

{
For Dataset-2, we do $10$-fold cross validation and choose the number of topics as $10$ from the Figure \ref{fig:cross2} below. In this case, we create the $10$ groups of documents in the following way. Out of $58$ documents, group $1$ has document number $1,11,21\ldots$, group $2$ has documents $2,22,32,\ldots$, etc. Sub sampling in this fashion respects the ordering in the documents. 
}

\begin{figure}
    \centering
    \includegraphics[scale = 0.35]{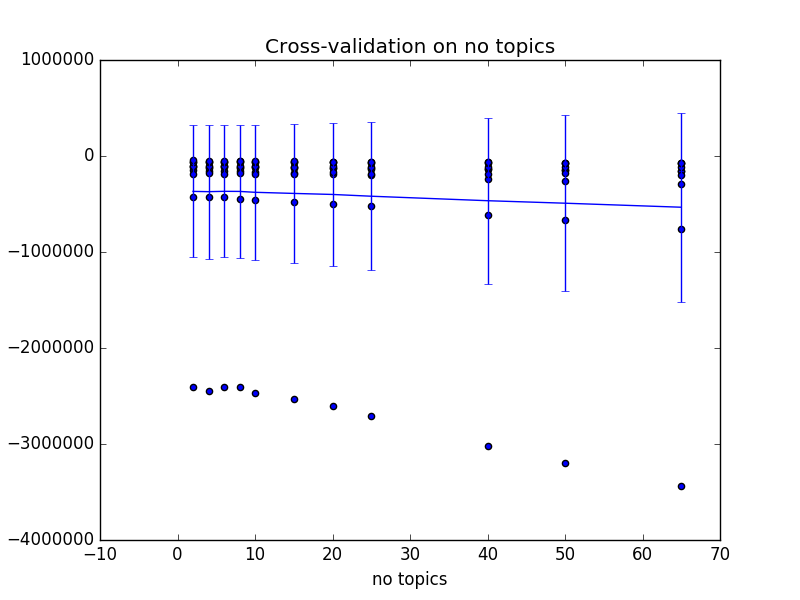}
    \caption{Likelihood vs number of topics in Dataset-2}
    \label{fig:cross2}
\end{figure}
\section{Conclusions and future work}
In this paper we look at the problem of detecting events in an error log generated by a distributed data center network. The error log consists of error messages with time stamps. Our goal is to detect latent events which 
generate these messages and find the distribution of messages for each event. We solve this problem by relating it to the topic modelling problem in documents. We introduce a notion of episodes in the time series data which serves as the equivalent of documents. Also we propose a linear time change detection algorithm to detect these episodes. We present consistency and sample complexity results for this change detection algorithm. Further we demonstrate the performance of our algorithm on a real dataset by comparing it with two benchmark algorithms existing in the literature.
We believe, our approach is
generic enough to be applied to other problem settings where the data has similar characteristics as network logs.
\bibliographystyle{IEEEtran}
\bibliography{myref,ref}
\clearpage
\section{SUPPLEMENTARY MATERIAL}
This is the supplementary material for the paper `Learning Latent Events from Network Message Logs'. We are making a code of our algorithm available at \url{https://github.com/siddpiku/CD-LDA}. The code also also includes the generation of a synthetic dataset on which one can run the algorithm.
\appendices

\section{Which algorithm for inference in LDA model?}
    \label{app:ldamodel}
 In order to perform this inference of event and episode signatures using topic modelling, many inference techniques exist: Gibbs sampling on the LDA model\cite{gibbs}, variational inference \cite{LDA}, online variational inference \cite{vr1}, stochastic variational inference \cite{vr2}. There are also provable inference models based on spectral methods, such as the tensor decomposition method in \cite{anima} and the SVD based method in \cite{kannan}. We use one of the popular python package based on Gibbs sampling based inference, \cite{gibbs}, for the real data experiments. One can also choose to use other more recent methods for inference as mentioned above.
    We work in the region where the number of messages are much larger that than the number of types of messages. In this region we show that most of the inference algorithms perform the same for our problem through a synthetic data experiment.
  
    So we compare three different inference algorithms, namely, Gibbs sampling on the LDA model\cite{gibbs}, online variational inference \cite{vr1} and the tensor decomposition method in \cite{anima}. We build an example with $4$ types of messages and $10000$ messages. We generate the time series as follows: There are two events, event $1$ has message distribution $[0.25,0.25,0.499,1e-3]$ and event $2$ has message distribution $[0.25,0.25,1e-3,0.499]$. Episode $1$ starts from message $1$ to message $3500$ and has only event $1$; episode $2$ starts from message $3501$ to $6054$ and has half of event $1$ and half of event $2$. Episode $3$ begins at message $6055$ and continues till the end with only event $2$ occurring in this episode. We run change detection based on $l_1$ metric followed by three different types of topic modeling inference algorithms on the episodes. We compare the inferred event-message distribution to the
    true event message distribution by computing the $l_1$ norm between the estimated and the true distribution maximized over all events. Table \ref{tab:LDA} summarizes the results. We can see that the error in estimating the event-message distributions are in the same order of magnitude.
    \begin{table}
    \begin{center}
    \begin{tabular}{|c|c|c|}                     
        \hline
        \multicolumn{3}{|c|}{ \shortstack{$l_1$ norm between the estimated and the true event-message distribution \\ maximized over all events} } \\
        \hline
                                                Gibbs Sampling, \cite{gibbs} &  Variational Inference, \cite{vr1} & Spectral LDA, \cite{anima}\\
        \hline
                                                0.014 & 0.021 & 0.094 \\
        \hline
    \end{tabular}
    \end{center}
    \caption{Comparison between inference methods for topic modelling}\label{tab:LDA}
    \end{table}

\section{Proof for multiple change point case, Theorem \ref{lemma:consistency}}
\label{app:multiple}
{To study the case of multiple change points, \cite{nonparamchpt} exploits the fact that their metric for change-point detection is convex between change points. 
However, the TV distance we use is not convex between two change points. But we work around this problem in the proof of theorem \ref{lemma:consistency} by showing that $D(\tg)$ is increasing to the left of the first change point, unimodal/increasing/decreasing between any two change points and decreasing to the right of the last change point. Hence, any global maximum of  $D(\tg )$ for $ 0<\tg <1$ is located at a change point.}

{Suppose we have more than one change points. We plan to show that $D(\tg n)\rightarrow D(\tg)$ and  $D(\tg)$ is increasing to the left of the first change point, unimodal/increasing/decreasing between two consecutive change points and decreasing to the right of last change point. If this happens, then we can conclude that one of the global maximas of $D(\tg)$ occurs at a change point. Using similar techniques from the single change point case, it is easy to show that $D(\tg)$ is increasing to the left of first change point and decreasing to the right of last change point (The proof is left to the readers as an exercise). Hence, it remains to show that  $D(\tg)$ is unimodal/increasing/decreasing between two consecutive change points. Lemma \ref{lemma:unimodal} proves this result. The prove of lemma \ref{lemma:unimodal} is relegated to Appendix \ref{app:unimodal} in the supplementary material.}

\begin{lemma}
{ $D(\g) = \lim_{n\rightarrow \infty} \D(\g n)$ is unimodal or increasing or decreasing between two consecutive change points when there is more than one change point.}
\label{lemma:unimodal}
\end{lemma}
\begin{remark}
{When we say $D(\g)$ is unimodal between two consecutive change points $\g_1<\g_2$, it means that there exists $\tg,$ $\g_1 < \tg < \g_2 $ such that $D'(\g) < 0$ for $\g<\tg$ and $D'(\g) > 0$ for $\g>\tg$.}
\end{remark}

\section{Proof of Lemma 4}
\label{app:unimodal}
Consider any two consecutive change points at index $\tau_1 = \g_1 n$ and $\tau_2 = \g_2 n$. Suppose the data points $X$ are drawn i.i.d from distribution $G$ between change points  $\tau_1$ and $\tau_2$. The data points to the left of $\tau_1$ are possibly drawn independently from more than one distribution. But, for the asymptotic analysis we can assume that the data points to the left of $\tau_1$ are possibly drawn i.i.d from the mixture of more than one distribution distribution. Lets call this mixture distribution $F$. Similarly, the data points to the right of $\tau_2$ can be assumed to be drawn i.i.d from a mixture distribution $H$. 
Let the inter-arrival time $\Delta t$ be drawn from  a distribution $F_t$ to the left of $\tau_1$ be, $G_t$ between $\tau_1$ and $\tau_2$ and $H_t$ to the right of $\tau_2$.

Suppose we consider the region $\tg$ between change points $\g_1$ and $\g_2$. 
So $\widehat{p}_L(\tg n )$ is a mixture of $\frac{\g_1}{\tg}$ fraction of samples from $F$ and $\frac{\tg - \g_1}{\tg}$ fraction from $G$.
$\widehat{p}_R(\tg n)$ is a mixture of $\frac{\g_{2} - \tg}{1 - \tg}$ fraction from $G$ and $\frac{1 - \g_{2}}{1 - \tg}$ fraction from $H$. So
\begin{align}
    \widehat{p}_L(\tg n )&\rightarrow\frac{\g_1}{\tg}F +  \frac{\tg - \g_1}{\tg}G\nonumber\\
    \widehat{p}_R(\tg n)&\rightarrow\frac{\g_{2} - \tg}{1 - \tg}G + \frac{1 - \g_{2}}{1 - \tg}H\label{eq:mixdist}
\end{align}

Similarly, the mean inter-arrival time of samples to the left of $\tg n$ converges to $\frac{\g_1}{\tg}\E F_t +  \frac{\tg - \g_1}{\tg}\E G_t$, and the mean inter-arrival time to the right of $\tg n$ converges to $\frac{\g_{2} - \tg}{1 - \tg}\E G_t + \frac{1 - \g_{2}}{1 - \tg}\E H_t$.
Combining this with \eqref{eq:mixdist}, we can say that 
\begin{align}
    \D(\tg n) \rightarrow D(\tg ) =   &\|\frac{\g_1}{\tg}(F - G) + \frac{1 - \g_{2}}{1 - \tg}(G-H)\|_1 \nonumber\\
    &+ |\frac{\g_1}{\tg}\E (F_t - G_t) + \frac{1 - \g_{2}}{1 - \tg}\E(G_t - H_t)|\label{eq:DLLN}
\end{align}
If we expand $\|\frac{\g_1}{\tg}(F - G) + \frac{1 - \g_{2}}{1 - \tg}(G-H)\|_1$ to sum of probabilities of individual messages as $\sum_{m=1}^M\frac{\g_1}{\tg}(F_m - G_m) + \frac{1 - \g_{2}}{1 - \tg}(G_m-H_m)$, we can write $D(\tg )$ from \eqref{eq:DLLN} as a function of $\tg$ as 
\begin{align}
D(\tg) &= \sum_{i=1}^{M+1} |\frac{a_i}{\tg} + \frac{b_i}{1 - \tg}|  \label{eq1}
\end{align}
for some constants $a_i,b_i\in \mathbb{R},i\in \{1,2,\ldots, M+1\}$.
Function $D(\tg)$ from \eqref{eq1} is only well defined over $\g_1<\tg<\g_2$. For the purpose of this proof, with some abuse of notation we assume the function $D(\g)$ to have the same definition outside $[\g_1,\g_2]$. We then show that $D(\tg)$ defined in \eqref{eq1} is unimodal/increasing/decreasing as a function of $\tg$ between $(0,1)$. This would naturally imply that $D(\tg)$ is unimodal/increasing/decreasing in $[\g_1,\g_2]$. The rest of the proof deals with this analysis.

Without loss of generality we can assume $a_i\ge 0,\;\;\forall i$. Note that $a_i,b_i\neq 0$ for all $i$. We can expand \eqref{eq1} as
\begin{align}
D(\tg) &= \sum_{a_i>0,b_i>0 } |\frac{a_i}{\tg} + \frac{b_i}{1-\tg}| + \sum_{a_i>0,b_i<0 } |\frac{a_i}{\tg} - \frac{-b_i}{1-\tg}|\nonumber
\\
&+ \sum_{b_i=0} \frac{a_i}{\tg}+\sum_{a_i=0} \frac{|b_i|}{1-\tg}\\
& = \frac{a}{\tg} + \frac{b}{1-\tg}+ \sum_{a_i,d_i>0 } |\frac{a_i}{\tg} - \frac{d_i}{1-\tg}|,
\label{eq2}
\end{align}
where $d_i = -b_i$ when $b_i<0$, $\sum_{i:b_i\ge 0} a_i= a$ and $\sum_{i:b_i>0} + b_i \sum_{a_i=0} |b_i| = b$.
$\frac{a_i}{\tg} - \frac{d_i}{1-\tg} > 0$ for $\tg<\frac{a_i}{a_i + d_i}$. We can assume w.l.o.g. that $\frac{a_i}{a_i + d_i}$ are in increasing order.
Suppose $\frac{a_{s}}{a_{s} + d_{s}} < \tg < \frac{a_{s+1}}{a_{s+1} + d_{s+1}}$.

\[D(\tg) =  \frac{a -\sum_{i<s} a_i + \sum_{i\ge s} a_i}{\tg} +  \frac{ b + \sum_{i<s} d_i - \sum_{i\ge s} d_i}{1-\tg}\]

Let $a(s) = a -\sum_{i<s} a_i + \sum_{i\ge s} a_i$ and $b(s) = b + \sum_{i<s} d_i - \sum_{i\ge s} d_i$. So for $\frac{a_{s}}{a_{s} + b_{s}} < \tg < \frac{a_{s+1}}{a_{s+1} + b_{s+1}}$
\begin{align}
    D(\tg) &=  \frac{a(s)}{\tg} + \frac{b(s)}{1-\tg},\;\; \frac{a_{s}}{a_{s} + b_{s}} < \tg < \frac{a_{s+1}}{a_{s+1} + b_{s+1}}\label{eq:Ds}
\end{align}

$a(s)>0$ for $s=0$  and it is a \textit{decreasing} function of $s$. $b(s)$ is a \textit{increasing} function of $s$. Based on where $a(s)$ changes sign w.r.t $b(s)$ we have the following cases. Note that $a(s)$ and $b(s)$ cannot both be negative for any value of $s$. $D'(\tg)$ denotes the derivative of $D(\tg)$ whereever it is   defined.
\begin{itemize}
    \item $a(s)>0,b(s)>0$ for all values of $s$.  $D(\tg )$ is a convex function of $\tg$ and hence is unimodal.
    \item $a(s)>0$ for all values of $s$ and $b(s)$ changes sign at $s = u$, i.e., $b(u)<0,b(u+1)>0$. So for $s\le u$, $D'(\tg)<0$ and for $s>u$, $D'(\tg )>0$. Hence, $D(\tg)$ is a unimodal function of $\tg $ between $0$ and $1$.
    \item $a(s)$ changes sign at $s = t$, i.e., $a(t)\ge 0, a(t+1)<0$ and $b(s)>0$ for all $s$. So for $s\le t$, $D(\tg n)$ is convex, and for $s> t $, $D'(\tg n)$ is positive. Hence, $D(\tg n)$ is either increasing or unimodal between $0$ and $1$.
    \item $a(t)\ge 0, a(t+1)<0$ and $b(u)\le 0,b(u+1)>0$. Also $t<u$. So for $s\le t$ $D'(\tg n)$ is decreasing, for $t<s\le u$ $D(\tg n)$ is convex and for $s>u$ $D'(\tg n)$ is increasing. Hence $D(\tg n)$ is unimodal.
\end{itemize}

\section{Proof of theorem \ref{th:multiplechptSmaple}}
\label{sec:multChPt}
\textbf{Proof for multiple change point case:}
Similar to the single change point case we first characterize the estimated change points $\hg_1,\ldots, \hg_k$ for finite $n$ in  lemma \ref{lemma:conc4}-\ref{lemma:conc6} below.

\begin{lemma}
$|\D(\tg n) - D(\tg )|\le \epsilon$ w.p. at least $ 1 - $ \\$4n \exp\left(  -\frac{\epsilon^2\alpha^2}{32\max(\sigma,k+1)^2}n + M\log(n + k)   \right)$ for all values of $\tg$.\label{lemma:conc4}
\end{lemma}
Lemma \ref{lemma:conc5} below can be proved in a similar way to lemma \ref{lemma:conc2} in single change point case.
\begin{lemma}
$|D(\g_i) - D(\hg_i)|<2\epsilon$ w.p. at least $ 1 -$\\
$4n \exp\left(  -\frac{\epsilon^2\alpha^2}{32\max(\sigma,k+1)^2}n + M\log(n + k)\right)$ for any change point $\g_i$.
\label{lemma:conc5}
\end{lemma}

\begin{lemma}
$ |\hg_i - \g_i|<c\epsilon$ w.p. at least $ 1 -$\\ $4n \exp\left(  -\frac{\epsilon^2\alpha^2}{32\max(\sigma,k+1)^2}n + M\log(n + k)\right)$ for some constant $c>0$ and any change point $\g_i$.\label{lemma:conc6}
\end{lemma}
Now, we can state the correctness result for Algorithm \ref{alg:multiplechangept}.
Algorithm \ref{alg:multiplechangept} is correct given accuracy $\epsilon>0$ as mentioned in Definition \ref{def:correctmultiple} with probability $1 - 7(2k+1)\exp(-\frac{(D^* - \epsilon)^2\epsilon^2\alpha^4}{512\max(\sigma^2,k+1)}n + M\log(n))$.

We upper bound the probability that Algorithm \ref{alg:multiplechangept} is not correct. From definition \ref{def:correctmultiple}, this happens when

\begin{itemize}
    \item Algorithm \ref{alg:multiplechangept} is correct every time it calls Algorithm \ref{alg:onechangept}.
\end{itemize}
The maximum number of times Algorithm \ref{alg:onechangept} would be applied is $2k+1$ if it is correct every time it is applied. Out of the $2k+1$ times $k$ number of times should return a change point and $k+1$ number of times should return no change point.
So
\begin{align}
    &P(\text{Algorithm \ref{alg:multiplechangept} is NOT correct})\nonumber
    \\
    &\le k P(\text{Algorithm \ref{alg:onechangept} does NOT detect a change point}\nonumber\\
    &\text{when one exists for data-set } X_L,\ldots,X_H)\nonumber
    \\
    &+ (k+1) P( \text{Algorithm \ref{alg:onechangept} returns a change point}\nonumber\\
    &\text{when one does not exist})
\end{align}
We assume that $X_L,\ldots,X_H$ is at least of size $\alpha n$ or an episode is at least $\alpha n$ samples long. Let $D^*$ denote the minimum value of metric $D$ at a global maxima for the reduced problem of $X_L,\ldots,X_H$ over all possible values of $L,H$ for which Algorithm \ref{alg:onechangept} is applied. From the correctness result for one change point, we have that

\begin{align}
    &P(\text{Algorithm \ref{alg:onechangept} does NOT detect a change point}\nonumber\\
    &\text{when one exists for data-set } X_L,\ldots,X_H)\nonumber
    \\
    &\le 3n\exp\left(  -\frac{\epsilon^2\alpha^2}{128\sigma^2}n + M\log(n)   \right)\nonumber\\
    &+ 3n\exp\left(  -\frac{\epsilon^2D^*}{512\sigma^2}n + M\log(n)   \right).
\end{align}
and
\begin{align}
    &P( \text{Algorithm \ref{alg:onechangept} returns a change point}\nonumber\\
    &\text{when one does not exist})\nonumber
    \\
    &\le (n+2)^V\exp(- n\delta^2/8).
\end{align}

Combining the above two cases, we get
\begin{align}
    &P(\text{Algorithm \ref{alg:multiplechangept} is NOT correct}) \nonumber\\
    &\le 7(2k+1)\exp(-\frac{(D^* - \epsilon)^2\epsilon^2\alpha^4}{512\max(\sigma^2,k+1)}n + M\log(n))\label{eq:samplemultiple}
\end{align}
Finally, under the assumptions
\begin{itemize}
    \item $\epsilon < \frac{D^*}{2}$,
    \item $\alpha +\epsilon < \min_r |\g_r -\g_{r-1}|$,
    \item $\delta < D^* - \epsilon$.
\end{itemize}
we can derive the sample complexity result for $k$  change points from \eqref{eq:samplemultiple}.

\section{Proof of Lemma \ref{lemma:conc1}}
\label{app:l2}
For notational simplicity, for any $\tg$, suppose $\widehat{p}_L(\tg n)\rightarrow p_L(\tg)$ and $\widehat{p}_R(\tg n)\rightarrow p_R(\tg)$

Since $|\D(\tg n) - D(\tg)| \le |\;\;\|\widehat{p}_L(\tg n)  - \widehat{p}_R(\tg n)\| - \|p_L(\tg n) - q(\tg n)\| \;\;| + |\;\;|\widehat{\E} S_1(\tg n) - \widehat{\E} S_2(\tg n)| - |\widehat{\E} S_1(\tg n) - \widehat{\E} S_2(\tg n)|\;\;|$,
\begin{align*}
  &P(|\D(\tg n) - D(\tg n)|>\epsilon)
  \\
  &\le P(|\;\;\|\widehat{p}(\tg n)  - \widehat{q}(\tg n)\| - \|p(\tg n) - q(\tg n)\| \;\;| > \frac{\epsilon}{2} ) \\
  &+ P(|\;\;|\widehat{m}_1(\tg n) - \widehat{m}_2(\tg n)| - |m_1(\tg n) - m_2(\tg n)|\;\;| > \frac{\epsilon}{2})\label{eq:conc} 
\end{align*}
First we focus on finding an upper bound to the probability $P(|\;\;\|\widehat{p}(\tg n)  - \widehat{q}(\tg n)\| - \|p(\tg n) - q(\tg n)\| \;\;| > \frac{\epsilon}{2} )$. 
\begin{align}
    &P(|\;\;\|\widehat{p}(\tg n)  - \widehat{q}(\tg n)\| - \|p(\tg n) - q(\tg n)\| \;\;| > \frac{\epsilon}{2} )
    \\
    &\le P(\|\widehat{p}(\tg n)  - \widehat{q}(\tg n)\|  > \|p(\tg n) - q(\tg n)\| + \frac{\epsilon}{2} )\nonumber\\
    &+ P(\|\widehat{p}(\tg n)  - \widehat{q}(\tg n)\|  < \|p(\tg n) - q(\tg n)\| - \frac{\epsilon}{2} ).\label{eq:pq}
\end{align}
Using $\|\widehat{p}(\tg n)  - \widehat{q}(\tg n)\|\le \|\widehat{p}(\tg n) - p(\tg n)\| + \|\widehat{q}(\tg n) - q(\tg n)\| + \|p(\tg n) - q(\tg n)\|$ and
$\|\widehat{p}(\tg n)  - \widehat{q}(\tg n)\| \ge \|p(\tg n) - q(\tg n)\| - \|\widehat{p}(\tg n) - p(\tg n)\| - \|\widehat{q}(\tg n) - q(\tg n)\|$ in \eqref{eq:pq} we have,

\begin{align}
    &P(|\;\;\|\widehat{p}(\tg n)  - \widehat{q}(\tg n)\| - \|p(\tg n) - q(\tg n)\| \;\;| > \frac{\epsilon}{2} )
    \\
    &\le 2P(\|\widehat{p}(\tg n) - p(\tg n)\| + \|\widehat{q}(\tg n) - q(\tg n)\|>\frac{\epsilon}{2})
    \\
    &\le 2P(\|\widehat{p}(\tg n) - p(\tg n)\| > \frac{\epsilon}{4}) + 2P(\|\widehat{q}(\tg n) - q(\tg n)\|>\frac{\epsilon}{4})\label{eq:my}
\end{align}

Let $\tg>\g$.  Now, $\|\widehat{p}(\tg n) - p(\tg n)\| \le \frac{\g}{\tg}\| \widehat{p}(\g n) - p\| + \frac{\tg - \g}{\tg}\|\widehat{p}(\g n:\tg n)  - q\|$. So,

\begin{align}
    &P(\|\widehat{p}(\tg n) - p(\tg n)\| > \frac{\epsilon}{4})
    \\
    &\le P(\frac{\g}{\tg} \| \widehat{p}(\g n) - p\| > \frac{\epsilon}{8}) + P(\frac{\tg - \g}{\tg}\|\widehat{p}(\g n:\tg n)  - q\| >\frac{\epsilon}{8})\label{eq:pconc}
\end{align}

We will apply Sanov's lemma to find an upper bound to $P(\frac{\g}{\tg} \| \widehat{p}(\g n) - p\| > \frac{\epsilon}{8})$. Consider the set $E$ of empirical probability distributions from i.i.d samples $X_1,\ldots,X_{\g n}$. $E = \{\widehat{p}(\g n): \frac{\g}{\tg} \| \widehat{p}(\g n) - p\| > \frac{\epsilon}{8}\}$. By Sanov's lemma we can say that,

\begin{align}
    P(E)&\le (\g n+1)^V\exp(-n \min_{p^* \in E} D_{KL}(p^*||p))\label{eq:sanov}
\end{align}

Further, by Pinsker's inequality, we have $D_{KL}(p^*||p)\ge \frac{1}{2} \|p^* - p\|^2$. Using this in \eqref{eq:sanov},

\begin{align}
    P(\frac{\g}{\tg} \| \widehat{p}(\g n) - p\| > \frac{\epsilon}{8})&\le (\g n+1)^V\exp\left(-\frac{n\epsilon^2}{128}\left(\frac{\tg}{\g}\right)^2\right)\label{eq:part1}
\end{align}

A similar approach yields 

\begin{align}
    &P(\frac{\tg - \g}{\tg}\|\widehat{p}(\g n:\tg n)  - q\| >\frac{\epsilon}{8})\nonumber\\
    &\le ((\tg - \g)n + 1)^V\exp\left(-\frac{n\epsilon^2}{128}\left(\frac{\tg}{\tg - \g}\right)^2\right).\label{eq:part2}
\end{align}

Combining \eqref{eq:part1} and \eqref{eq:part2} and substituting in \eqref{eq:pconc}, we get

\begin{align}
    &P(\|\widehat{p}(\tg n) - p(\tg n)\| > \frac{\epsilon}{4})\le (\g n+1)^V\exp\left(-\frac{n\epsilon^2}{128}\left(\frac{\tg}{\g}\right)^2\right)\nonumber\\
    &+ ((\tg - \g)n + 1)^V\exp\left(-\frac{n\epsilon^2}{128}\left(\frac{\tg}{\tg - \g}\right)^2\right).\label{eq:part3}
\end{align}

Also using Sanov's theorem followed by Pinsker's inequality we have,
\begin{align}
    P(\|\widehat{q}(\tg n)  - q\| >\frac{\epsilon}{4})&\le ((1 - \tg)n + 1)^V\exp\left(-\frac{n\epsilon^2}{32}\right).\label{eq:part4}
\end{align}

Finally, \eqref{eq:part3} and \eqref{eq:part4} yield the following inequality using \eqref{eq:my},

\begin{align}
    &P(|\;\;\|\widehat{p}(\tg n)  - \widehat{q}(\tg n)\| - \|p(\tg n) - q(\tg n)\| \;\;| > \frac{\epsilon}{2} )
    \\
    &\le 2(\g n+1)^V\exp\left(-\frac{n\epsilon^2}{128}\left(\frac{\tg}{\g}\right)^2\right)\nonumber\\
    &+ 2((\tg - \g)n + 1)^V\exp\left(-\frac{n\epsilon^2}{128}\left(\frac{\tg}{\tg - \g}\right)^2\right) \\
    &+ 2((1 - \tg)n + 1)^V\exp\left(-\frac{n\epsilon^2}{32}\right)
    \\
    &\le 3 \exp\left(  -\frac{\epsilon^2\alpha^2}{128}n + V\log(n)   \right)\label{eq:pconc}
\end{align}
We can get a result same as \eqref{eq:pconc} for $\tg<\g$.

Now, lets prove concentration results for $g(\tg)$.
\[g(\tg) = \widehat{\E} S_1(\tg n) - \widehat{\E} S_2(\tg n) = \frac{\sum_{j=1}^{\tg n} \Delta t_j}{\tg n} - \frac{\sum_{j=\tg n+1}^{n} \Delta t_j}{(1-\tg)n}.\]

By assumption, $\Delta t_j$ is subgaussian from $j= 1$ to $\g n$ with parameter $\sigma_1^2$ and from $j= \g n +1$ to $\g n$ with parameter $\sigma_2^2$. If $\Delta t_j$ is subgaussian, so is r.v. $-\Delta t_j$ with the same subgaussian parameter. Sum of suggaussian r.v is also subgaussian with parameter equal to the sum of individual subgaussian parameters. Let $\sigma = \max (\sigma_1,\sigma_2)$.
So, the sum of subgaussian parameters for $g(\tg) $, say $\sigma_{g}$, is upper bounded by

\[\sigma_g^2 \le \sum_{j=1}^{\tg n} \frac{\sigma^2}{\tg^2 n^2} + \sum_{j=\tg n +1}^{n} \frac{\sigma^2}{(1 - \tg)^2n^2} \le \frac{\sigma^2}{\alpha^2n} \] 

\begin{align}
    P(|g(\tg n) - \E g(\tg n)|>\frac{\epsilon}{2})&\le 2\exp\left(-\frac{\epsilon^2}{8\sigma_g^2}\right)\\
    &\le 2\exp\left(-n\epsilon^2\frac{\alpha^2}{8\sigma^2}\right)\label{eq:gconc}
\end{align}

Putting together \eqref{eq:gconc} and \eqref{eq:pconc} with \eqref{eq:conc},

\begin{align}
    P(|\D(\tg n) - D(\tg n)|>\epsilon)
    &\le 3 \exp\left(  -\frac{\epsilon^2\alpha^2}{128}n + V\log(n)   \right) \nonumber\\
    &+ 2\exp\left(-n\epsilon^2\frac{\alpha^2}{8\sigma^2}\right)
    \\
    &\le 3 \exp\left(  -\frac{\epsilon^2\alpha^2}{128\sigma^2}n + V\log(n)   \right) 
\end{align}
For all values of $\tg$, we have by union bound,
\begin{align}
    &P(|\D(\tg n) - D(\tg n)|<\epsilon,\text{ for all } \alpha< \tg<1 - \alpha)\nonumber\\
    &\le 1 - 3n\exp\left(  -\frac{\epsilon^2\alpha^2}{128\sigma^2}n + V\log(n)   \right).
\end{align}
\section{Proof of Lemma \ref{lemma:conc2} and Lemma \ref{lemma:conc3}}
\label{app:l3l4}
From \eqref{Dlln} $\arg \max_{\tg} D(\tg n) = \g$. Let $\arg \max_{\tg} \D(\tg n) = \hg$.

Now, w.p. $ 1 - 3n\exp\left(  -\frac{\epsilon^2\alpha^2}{128\sigma^2}n + V\log(n)   \right)$

\begin{align*}
    D(\g n) - \D(\hg) &< \D(\g n) - \D(\hg n) + \epsilon<\epsilon 
\end{align*}
Also,
\begin{align*}
    D(\g n) - \D(\hg) &> D(\g n) - D(\hg n) - \epsilon> -\epsilon 
\end{align*}

So we have,

\[|D(\g n) - \D(\hg n)|\le \epsilon \] w.p. $1 - 3n\exp\left(  -\frac{\epsilon^2\alpha^2}{128\sigma^2}n + V\log(n)   \right)$.

Now
$0<D(\g n) - D(\hg n)<D(\g n) - \D(\hg n) + \epsilon < 2\epsilon$.

Suppose $\hg > \g$. So by \eqref{Dlln}, $|D(\g n) - D(\hg n)| = D(\g n)| \frac{\hg  - \g}{\hg}| > \frac{D(\g n)}{1 - \alpha} |\hg - \g|$.

  For $\hg < \g$, $|D(\g n) - D(\hg n)| = D(\g n)| \frac{\hg  - \g}{1 - \hg}| > \frac{D(\g n)}{1 - \alpha} |\hg - \g|$.

\section{Proof of Lemma \ref{lemma:conc4}}
Since $|\D(\tg n) - D(\tg n)| \le |\;\;\|\widehat{p}(\tg n)  - \widehat{q}(\tg n)\| - \|p(\tg n) - q(\tg n)\| \;\;| + |\;\;|\widehat{m}_1(\tg n) - \widehat{m}_2(\tg n)| - |m_1(\tg n) - m_2(\tg n )\;\;|$,
\begin{align*}
  &P(|\D(\tg n) - D(\tg n)|>\epsilon)
  \\
  &\le P(|\;\;\|\widehat{p}(\tg n)  - \widehat{q}(\tg n)\| - \|p(\tg n) - q(\tg n)\| \;\;| > \frac{\epsilon}{2} ) \nonumber\\
  &+ P(|\;\;|\widehat{m}_1(\tg n) - \widehat{m}_2(\tg n)| - |m_1(\tg n) - m_2(\tg n)|\;\;| > \frac{\epsilon}{2})\label{eq:conc} 
\end{align*}
First we focus on finding an upper bound to the probability $P(|\;\;\|\widehat{p}(\tg n)  - \widehat{q}(\tg n)\| - \|p(\tg n) - q(\tg n)\| \;\;| > \frac{\epsilon}{2} )$. 
\begin{align}
    &P(|\;\;\|\widehat{p}(\tg n)  - \widehat{q}(\tg n)\| - \|p(\tg n) - q(\tg n)\| \;\;| > \frac{\epsilon}{2} )
    \\
    &\le P(\|\widehat{p}(\tg n)  - \widehat{q}(\tg n)\|  > \|p(\tg n) - q(\tg n)\| + \frac{\epsilon}{2} )\nonumber\\
    &+ P(\|\widehat{p}(\tg n)  - \widehat{q}(\tg n)\|  < \|p(\tg n) - q(\tg n)\| - \frac{\epsilon}{2} ).\label{eq:pq}
\end{align}
Using 
$\|\widehat{p}(\tg n)  - \widehat{q}(\tg n)\|\le \|\widehat{p}(\tg n) - p(\tg n)\| + \|\widehat{q}(\tg n) - q(\tg n)\| + \|p(\tg n) - q(\tg n)\|$
and
$\|\widehat{p}(\tg n)  - \widehat{q}(\tg n)\| \ge \|p(\tg n) - q(\tg n)\| - \|\widehat{p}(\tg n) - p(\tg n)\| - \|\widehat{q}(\tg n) - q(\tg n)\|$

in \eqref{eq:pq}, we have,

\begin{align}
    &P(|\;\;\|\widehat{p}(\tg n)  - \widehat{q}(\tg n)\| - \|p(\tg n) - q(\tg n)\| \;\;| > \frac{\epsilon}{2} )
    \\
    &\le 2P(\|\widehat{p}(\tg n) - p(\tg n)\| + \|\widehat{q}(\tg n) - q(\tg n)\|>\frac{\epsilon}{2})
    \\
    &\le 2P(\|\widehat{p}(\tg n) - p(\tg n)\| > \frac{\epsilon}{4}) + 2P(\|\widehat{q}(\tg n) - q(\tg n)\|>\frac{\epsilon}{4})\label{eq:my}
\end{align}

Let $\g_r < \tg < \g_{r+1}$. Now, $\|\widehat{p}(\tg n) - p(\tg n)\| \le\sum_{j=0}^r \frac{\g_j -\g_{j-1}}{\tg}\| \widehat{p}^{j-1} - p^{j-1}\| + \frac{\tg -\g_{r}}{\tg} \|\widehat{p}(\g n:\tg n)  - p^{r}\|$. So,

\begin{align}
    &P(\|\widehat{p}(\tg n) - p(\tg n)\| > \frac{\epsilon}{4})
    \\
    &\le \sum_{j=0}^r P(\frac{\g_j - \g_{j-1}}{\tg} \| \widehat{p}^{j-1} - p^{j-1}\| > \frac{\epsilon}{4(r+1)}) \nonumber\\
    &+ P(\frac{\tg - \g_r}{\tg}\|\widehat{p}(\g_r n:\tg n)  - p^{r}\| >\frac{\epsilon}{4(r+1)})\label{eq:pconc}
\end{align}

We will apply Sanov's theorem to find an upper bound to 
$P(\frac{\g_j - \g_{j-1}}{\tg} \| \widehat{p}^{j-1} - p^{j-1}\| > \frac{\epsilon}{4(r+1)})$.
Consider the set $E$ of empirical probability distributions from i.i.d samples $X_{\g_{j-1} n},\ldots,X_{\g_j n}$. $E = \{\widehat{p}^{j-1}: \frac{\g_j - \g_{j-1}}{\tg} \| \widehat{p}^{j-1} - p^{j-1}\| > \frac{\epsilon}{4(r+1)}\}$. By Sanov's theorem we can say that,

\begin{align}
    P(E)&\le ((\g_j - \g_{j-1}) n+1)^V\exp(-n \min_{p^* \in E} D_{KL}(p^*||p^{j-1}))\label{eq:sanov}
\end{align}

Further, by Pinsker's inequality, we have $D_{KL}(p^*||p^{j-1})\ge \frac{1}{2} \|p^* - p^{j-1}\|^2$. Using this in \eqref{eq:sanov},

\begin{align}
    &P(\frac{\g_j - \g_{j-1}}{\tg} \| \widehat{p}^{j-1} - p^{j-1}\| > \frac{\epsilon}{4(r+1)})\nonumber\\
    &\le ((\g_j - \g_{j-1}) n+1)^V\exp\left(-\frac{n\epsilon^2}{32(r+1)^2}\left(\frac{\tg}{\g_j - \g_{j-1}}\right)^2\right)\label{eq:part1}
\end{align}

A similar approach yields 

\begin{align}
    &P(\frac{\tg - \g_r}{\tg}\|\widehat{p}(\g n:\tg n)  - p^{j-1}\| >\frac{\epsilon}{4(r+1)})\nonumber\\
    &\le ((\tg - \g_r)n + 1)^V\exp\left(-\frac{n\epsilon^2}{32(r+1)^2}\left(\frac{\tg}{\tg - \g_r}\right)^2\right).\label{eq:part2}
\end{align}

Combining \eqref{eq:part1} and \eqref{eq:part2} and substituting in \eqref{eq:pconc}, we get
\begin{align}
    &P(\|\widehat{p}(\tg n) - p(\tg n)\| > \frac{\epsilon}{4})\nonumber\\
    &\le \sum_{j =0}^r((\g_j - \g_{j-1})n+1)^V\times\nonumber\\
    &\exp\left(-\frac{n\epsilon^2}{32(r+1)^2}\left(\frac{\tg}{\g_j - \g_{j-1}}\right)^2\right) \nonumber
    \\
    &+ ((\tg - \g_r)n + 1)^V\exp\left(-\frac{n\epsilon^2}{32(r+1)^2}\left(\frac{\tg}{\tg - \g_r}\right)^2\right)\nonumber
    \\
    &\le \left(\sum_{j =0}^r((\g_j - \g_{j-1})n+1)^V + ((\tg - \g_r)n + 1)^V\right)\times\nonumber\\
    &\exp\left(-\frac{n\epsilon^2\alpha^2}{32(k+1)^2}\right)\label{eq:part3}
\end{align}

Also using Sanov's theorem followed by Pinsker's inequality we have,
\begin{align}
    &P(\|\widehat{q}(\tg n) - q(\tg n)\| > \frac{\epsilon}{4})\nonumber\\
    &\le \left(\sum_{j =r+1}^k((\g_j - \g_{j-1})n+1)^V + ((\g_{r+1} - \tg)n + 1)^V\right)\times\nonumber\\
    &\exp\left(-\frac{n\epsilon^2\alpha^2}{32(k+1)^2}\right)\label{eq:part4}
\end{align}

Finally, \eqref{eq:part3} and \eqref{eq:part4} yield the following inequality using \eqref{eq:my},

\begin{align}
    &P(|\;\;\|\widehat{p}(\tg n)  - \widehat{q}(\tg n)\| - \|p(\tg n) - q(\tg n)\| \;\;| > \frac{\epsilon}{2} )\nonumber
    \\
    &\le 2(\sum_{j =0}^r((\g_j - \g_{j-1})n+1)^V + ((\tg - \g_r)n + 1)^V\nonumber\\
    &+ ((\g_{r+1} - \tg)n + 1)^V\nonumber\\
    &+ \sum_{j =r+1}^k((\g_j - \g_{j-1})n+1)^V  )\exp\left(-\frac{n\epsilon^2\alpha^2}{32(k+1)^2}\right)
    \\
    &\le 2 \exp\left(  -\frac{\epsilon^2\alpha^2}{32(k+1)^2}n + V\log(n +k)   \right)\label{eq:pconc}
\end{align}

Now, lets prove concentration results for $g(\tg)$.
\[g(\tg) = \widehat{\E} S_1(\tg n) - \widehat{\E} S_2(\tg n) = \frac{\sum_{j=1}^{\tg n} \Delta t_j}{\tg n} - \frac{\sum_{j=\tg n+1}^{n} \Delta t_j}{(1-\tg)n}.\]

By assumption, $\Delta t_j$ is subgaussian from $j= 1$ to $\g_1 n$ with parameter $\sigma_1^2$ and from $j= \g_1 n +1$ to $\g_2 n$ with parameter $\sigma_2^2$ and so on. If $\Delta t_j$ is subgaussian, so is r.v. $-\Delta t_j$ with the same subgaussian parameter. Sum of suggaussian r.v is also subgaussian with parameter equal to the sum of individual subgaussian parameters. Let $\sigma = \max (\sigma_1,\sigma_2,\ldots,\sigma_k)$.
So, the sum of subgaussian parameters for $g(\tg) $, say $\sigma_{g}$ , is upper bounded by

\[\sigma_g^2 \le \sum_{j=1}^{\tg n} \frac{\sigma^2}{\tg^2 n^2} + \sum_{j=\tg n +1}^{n} \frac{\sigma^2}{(1 - \tg)^2n^2} \le \frac{\sigma^2}{\alpha^2n} \] 

\begin{align}
    P(|g(\tg n) - \E g(\tg n)|>\frac{\epsilon}{2})&\le 2\exp\left(-\frac{\epsilon^2}{8\sigma_g^2}\right)\nonumber\\
    &\le 2\exp\left(-n\epsilon^2\frac{\alpha^2}{8\sigma^2}\right)\label{eq:gconc}
\end{align}

Putting together \eqref{eq:gconc} and \eqref{eq:pconc} with \eqref{eq:conc},

\begin{align}
    &P(|\D(\tg n) - D(\tg n)|>\epsilon)\nonumber\\
    &\le 2 \exp\left(  -\frac{\epsilon^2\alpha^2}{32(k+1)^2}n + V\log(n +k)\right)\nonumber\\
    &+ 2\exp\left(-n\epsilon^2\frac{\alpha^2}{8\sigma^2}\right)\nonumber
    \\
    &\le 4 \exp\left(  -\frac{\epsilon^2\alpha^2}{32\max(\sigma,k+1)^2}n + V\log(n + k)   \right) 
\end{align}
For all values of $\tg$, we have by union bound,
\begin{align}
    &P(|\D(\tg n) - D(\tg n)|<\epsilon,\text{ for all } \alpha< \tg<1 - \alpha)\nonumber\\
    &\le 1 - 4n \exp\left(  -\frac{\epsilon^2\alpha^2}{32\max(\sigma,k+1)^2}n + V\log(n + k)   \right)  .
\end{align}
\section{Proof of Lemma \ref{lemma:conc6}}
From Lemma \ref{lemma:conc5}, 
$|D(\gamma n) - D(\hg n)|<2\epsilon$ w.p. at least $ 1 -$ \\$ 4n \exp\left(  -\frac{\epsilon^2\alpha^2}{32\max(\sigma,k+1)^2}n + V\log(n + k)   \right)$. Also, from lemma \ref{lemma:unimodal} we know that $\g n$ is a change point, and all local maximas in $D(\tg n)$ for $0<\tg<1$ correspond to a change point. Suppose that $\g_r$ is a change point closest to $\hg$ such that $|D(\g_r n) - D(\hg n)|<2\epsilon$. Also, since $D(\tg n)$ for $0<\tg<1$ is unimodal or monotonic between $\g_r,\g_{r+1}$ or $\g_{r-1},\g_r$ we assume w.l.o.g that $D'(\g_r n)$
and $D(\hg n)$ have the same sign. Hence, $D(\tg n)$, $\tg$ between $\g_r$ and $\hg$ is monotonic. We want to lower bound $\frac{|D(\g_r n) - D(\hg n)|}{|\g_r n - \hg n|}$. W.l.o.g we assume that $\g_r < \hg$ and $D(\tg n)$, $\tg$ between $\g_r$ and $\hg$, is decreasing.

From \eqref{eq2} we know the expression for $D(\tg n)$  as 
\begin{align}
D(\tg n) &= \frac{a}{\tg} + \frac{b}{1 - \tg}
\end{align}
for some constants $a,b$. The constants $a$ and $b$ may change over different ranges of $\tg$ between $\g_r$ and $\hg$.
Consider a range of $\tg$ between $\g_r$ and $\hg$ over which $a,b$ are constant. Now, consider the difference $D(\g^1 n) - D(\g^2 n)$ for $\g^1<\g^2$ belonging to that range. We will lower bound $\frac{D(\g^1 n) - D(\g^2 n)}{\g^2 n - \g^1 n}$ for different values of $a,b$. 

\begin{itemize}
    \item $a>0,b>0$. So,
    \begin{align}
        \frac{D(\g^1 n) - D(\g^2 n)}{(\g^2 - \g^1)} &= \left(\frac{a}{\g^1\g^2} - \frac{b}{(1 - \g^2)(1 - \g^1)}\right) \label{eq:gammadec}
    \end{align}
    Now \eqref{eq:gammadec} is a decreasing function of $\g^2$ since $a,b>0$. Now $\frac{D(\g^1 n) - D(\g^2 n)}{(\g^2 - \g^1)}$ is a minimum when $\g^2 - \g^1$ is maximum. $\g^2 - \g^1$ is maximum when $D(\g^1 n) - D(\g^2 n)$ is $2\epsilon$.  So, $\frac{D(\g^1 n) - D(\g^2 n)}{(\g^2 - \g^1)}$ is $c(\epsilon,a,b)>0$ at minimum, where $c$ is some constant as a function of $2\epsilon,a,b$ .
    \item $a>0,b<0$.
    
    \begin{align}
        \frac{D(\g^1 n) - D(\g^2 n)}{(\g^2 - \g^1)} &= \left(\frac{a}{\g^1\g^2} - \frac{b}{(1 - \g^2)(1 - \g^1)}\right)
        \\
        &\ge \left(\frac{a}{\g^1\g^2} + \frac{b}{(1 - \g^2)(1 - \g^1)}\right)
        \\
        &\ge \left(\frac{a}{\g^1} + \frac{b}{(1 - \g^1)}\right)
        \\
        & = D(\g^1 n)
        \label{eq:gammadec2}
    \end{align}
\end{itemize}

From the above two cases we can conclude that $\frac{D(\g^1 n) - D(\g^2 n)}{(\g^2 - \g^1)}\ge \min(D(\g^1 n),2\epsilon )$.

Suppose $a,b$ change values at $l$ different places between $\g_r$ and $\hg$. Let the points be denoted as $\g^1,\g^2, \ldots, \g^l$. So,

\begin{align}
    &D(\g_r n) - D(\hg n) \\
    &= D(\g_r n) - D(\g^1 n) + D(\g^1 n) - D(\g^2 n)\\
    &+ \ldots + D(\g^l n) - D(\hg n)
    \\
    &\ge \min(D(\g_r n),c(\epsilon,a^1,b^1)) (\g^1 -\g_r) +\ldots\nonumber\\
    &+ \min(D(\g_l n),c(\epsilon,a^l,b^l) ) (\hg -\g^l)
    \\
    &\ge \min(D(\g^l n),\min_{1<i<l} c(\epsilon,a_i,b_i) ) (\hg -\g_r)
    \\
    &\ge \min(D(\g_r n) - 2\epsilon,\min_{1<i<l} c(\epsilon,a_i,b_i) ) (\hg -\g_r)
\end{align}

So, 
\[(\hg -\g_r) \le \frac{2\epsilon}{\min(D(\g_r n) - 2\epsilon,\min_{1<i<l} c(\epsilon,a_i,b_i) )}\]

We can prove similarly when $\hg<\g^r$.
\section{Setup and Methodology for experiments}
\label{app:preprocessing}

{\bf Template extraction:}  Raw syslog data has three fields: timestamp, router id, and message text. Since the number of distinct messages are very large and many of them have common patterns, it is often useful~\cite{Makanju:2009:CEL:1557019.1557154,Li:2005:IFM:1081870.1081972,DBLP:conf/icdm/TangL10,Li:2017:DTC:3101309.3092697} to decompose the message text into two parts: an {\em invariant} part called template, and {\em parameters} associated with  template. For example, two different messages in the log can look like: 
\begin{itemize}[leftmargin=15pt]
\footnotesize
\item \texttt{Base SVCMGR-MINOR-sapCemPacketDefectAlarmClear-2212 [CEM SAP Packet Errors]: SAP 124 in service wqffv (customer 1): Alarm \'bfrUnderrun \' Port 23.334 Alarm \'bfrUnderrun \' 22333242 ,22595400}
\item \texttt{Base SVCMGR-MINOR-sapCemPacketDefectAlarmClear-2212 [CEM SAP Packet Errors]: SAP 231 in service qaazxs (customer 1): Alarm \'bfrUnderrun \' Port 3322 Alarm \'bfrUnderrun \' 22121222 ,22595400}
\end{itemize}
Ideally, we wish to extract the following template from these identical messages:
\begin{itemize}[leftmargin=15pt]
\footnotesize
	\item \texttt{Base SVCMGR-MINOR-sapCemPacketDefectAlarmClear-2212 [CEM SAP Packet Errors]: SAP * in service * (customer 1): Alarm \'bfrUnderrun \' Port * Alarm \'bfrUnderrun \' * ,22595400}
\end{itemize}
There are many existing methods to extract such templates\cite{lishwartz:kdd2017:tutorial, Li:2017:DTC:3101309.3092697}, ranging from tree-based methods to NLP based methods. In our work, we use an NLP based method as follows: (i) We compute the bigram probability of each word in the message corpus, (ii) next,  each words above a predetermined empirical probability is declared as a word belonging to a template, (iii) each message is converted into a template by substituting the non-template-words with~* as in the preceding paragraph, and  (iv) finally, we assign an id to each template-router tuple in every log entry. The last step essentially combines two fields in syslog, namely text message converted to template, and source/router field. The output of this last step is treated as {\em message} by CD-LDA and the other algorithms.  When we applied this steps to our first data set, we extracted 39,330 distinct template-router combinations.

Note that, when alarms are reported, the template extraction stage is redundant.
 
{\bf Additional pre-processing:}  Since each  event in a real-system has effects that last for several minutes to hours (even days at times), we are only interested in events at the time-scale of several minutes to an hour. Thus,  in this step, we round the time-steps from $msec$ granularity to minutes (or fraction of minute) . This temporal rounding helps us to speed-up our algorithms while serving the intended practical benefit. We chose 1~minute rounding for dataset-1 and 5~minute rounding for dataset-2. Note that, upon performing temporal rounding, we do not discard duplicate messages that could result from the rounding.

\section{The metric used in \cite{nonparamchpt}}
\label{app:L2unbiased}
In \cite{nonparamchpt} the data points lie in a continuous real space. We can still apply it to categorical data like ours if we encode a categorical data point $i\in\{1,2,\ldots,M\}$ as a vector with all zeros except for the location $i$.  If we use this encoding, we can show that the metric used in \cite{nonparamchpt} degenerates to an unbiased estimator of the squared $l_2$ norm. This encoding also helps us compute the metric in linear cost as oppposed to quadratic computation cost in \cite{nonparamchpt}.  The proof follows below. 

Suppose $X_1,\ldots,X_n$ are drawn i.i.d from $p$ and $Y_1,\ldots,Y_m$ are drawn i.i.d from $q$. Then  \cite{nonparamchpt} computes the similarity in the two distributions as,
\begin{align}
    &\widehat{E}(X,Y,\alpha) = \frac{2}{mn} \sum_{i,j} |X_i - Y_j|^\alpha \text{ ($\alpha\in (0,2)$) }\nonumber \\
    &\nonumber - {n \choose 2}^{-1} \sum_{i<j} |X_i - X_j|^\alpha - {m \choose 2}^{-1} \sum_{i<j} |Y_i - Y_j|^\alpha\\
    &\nonumber = \frac{2}{mn} \sum_{i,j} \ind{X_i\neq Y_j}  \\
    &- {n \choose 2}^{-1} \sum_{i<j} \ind{X_i\neq X_j} - {m \choose 2}^{-1} \sum_{i<j} \ind{Y_i\neq Y_j}\label{eq:l2metric}
\end{align}
Let $n_i$ denote the number of data points in $X_1,\ldots,X_n$ taking the value $i$ and $m_i$ denote the number of data points in $Y_1,\ldots,Y_m$ taking value $i$. One can reduce \eqref{eq:l2metric} to 
\begin{align}
    \widehat{E}(X,Y,\alpha) &= \sum_i \frac{n_i^2 - n_i}{n^2-n} + \frac{m_i^2 - m_i}{m^2-m} - 2\frac{n_im_i}{nm}\label{eq:l2unbiased} 
\end{align}

As $n,m\rightarrow \infty$, $\widehat{E}(X,Y,\alpha) \rightarrow \|p-q\|^2_2$. Also, $\E \widehat{E}(X,Y,\alpha) = \|p - q\|_2^2$. So $\widehat{E}(X,Y,\alpha)$ is both a consistent and unbiased estimator for $\|p - q\|_2^2$. 

\end{document}